\documentclass[letterpaper]{article} 
\usepackage{aaai2026}  
\usepackage{times}  
\usepackage{helvet}  
\usepackage{courier}  
\usepackage[hyphens]{url}  
\usepackage{graphicx}
\usepackage{booktabs}
\usepackage{natbib}  
\usepackage{caption} 
\usepackage{amsmath}
\usepackage{amssymb}     
\usepackage{graphicx} 
\usepackage[linesnumbered,ruled,vlined]{algorithm2e}
\usepackage{multirow}
 \usepackage{colortbl}
\usepackage{float}

\usepackage{xr}

\usepackage{newfloat}
\usepackage{listings}
\DeclareCaptionStyle{ruled}{labelfont=normalfont,labelsep=colon,strut=off} 
\floatstyle{ruled}

\newfloat{listing}{tb}{lst}{}
\floatname{listing}{Listing}
\title{Simulating Distribution Dynamics: Liquid Temporal Feature Evolution 

for Single-Domain Generalized Object Detection}

\author{Zihao Zhang\textsuperscript{1}, Yang Li\textsuperscript{1}, Aming Wu\textsuperscript{2}\thanks{Corresponding author.}, Yahong Han\textsuperscript{1}\\
\textsuperscript{1}College of Intelligence and Computing, Tianjin University, Tianjin, China\\
\textsuperscript{2}School of Computer Science and Information Engineering, Hefei University of Technology, China\\
{\tt\small zhangzihao2490@tju.edu.cn, liyang1398@tju.edu.cn, amwu@hfut.edu.cn,  yahong@tju.edu.cn}
}

\usepackage{bibentry}

\begin{document}

\maketitle

\begin{abstract}
In this paper, we focus on Single-Domain Generalized Object Detection (Single-DGOD), aiming to transfer a detector trained on one source domain to multiple unknown domains.
Existing methods for Single-DGOD typically rely on discrete data augmentation or static perturbation methods to expand data diversity, thereby mitigating the lack of access to target domain data. However, in real-world scenarios such as changes in weather or lighting conditions, domain shifts often occur continuously and gradually. 
Discrete augmentations and static perturbations fail to effectively capture the dynamic variation of feature distributions, thereby limiting the model's ability to perceive fine-grained cross-domain differences.
To this end, we propose a new method, i.e., Liquid Temporal Feature Evolution, which simulates the progressive evolution of features from the source domain to simulated latent distributions by incorporating temporal modeling and liquid neural network–driven parameter adjustment. Specifically, we introduce controllable Gaussian noise injection and multi-scale Gaussian blurring to simulate initial feature perturbations, followed by temporal modeling and a liquid parameter adjustment mechanism to generate adaptive modulation parameters, enabling a smooth and continuous adaptation across domains.
By capturing progressive cross-domain feature evolution and dynamically regulating adaptation paths, our method bridges the source-unknown domain distribution gap, significantly boosting generalization and robustness to unseen shifts.
Significant performance improvements on the Diverse Weather dataset and Real-to-Art benchmark demonstrate the superiority of our method. Our code is available at \url{https://github.com/2490o/LTFE}.
\end{abstract}    

\section{Introduction}

\label{sec:intro}
Single-domain Generalized Object Detection \cite{C_Cap, Div, UFR} is a challenging yet crucial task in object detection, as it requires the model to adapt to domain shifts that were not encountered during training. The core challenges faced in Single-DGOD can be summarized in two main aspects: First, the unobservability of domain shifts \cite{domainshift} prevents models from eliminating the distribution differences between the source domain and unknown domains using traditional feature alignment methods \cite{wu2-domain, wu3-domainInstanr, wu1-universal}. Second, the diversity of unknown domain data distributions \cite{Div} makes it difficult for models trained on a single source domain distribution to generalize to multiple unknown domains with different data distributions \cite{wu2023discriminating}.

\begin{figure}[!t]
	\centering
	\includegraphics[width=3.35in]{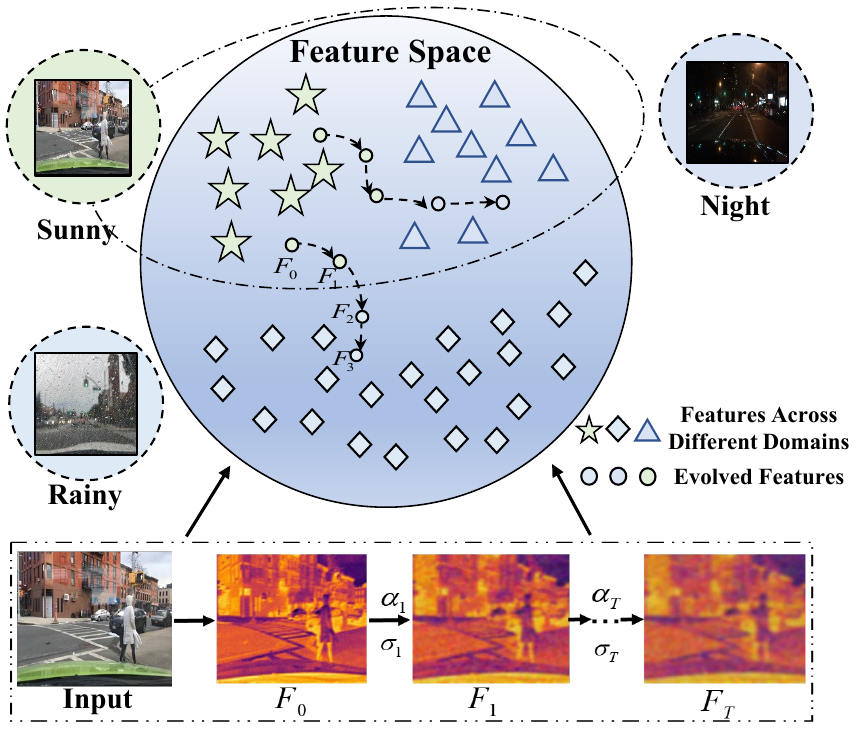}
	\DeclareGraphicsExtensions.
	\begin{center}
		\vspace{-5pt}
		\caption{Liquid Temporal Feature Evolution (LTFE) for detecting the unknown-domain object. The core of LTFE lies in simulating the feature evolution trajectory from the source domain to simulated latent distributions, capturing continuous cross-domain feature evolution. As illustrated, LTFE facilitates a smooth transition of features from the source domain to the simulated latent distributions.}
        \label{F1}
        \vspace{-10pt}
	\end{center}
	
\end{figure}
Existing Single-DGOD methods primarily enhance generalization to unknown domains through two approaches: one leverages the multimodal capabilities of vision-language models, using static textual prompts from the target domain to estimate domain shifts or simulate target styles \cite{C_Cap, Poda, pdoc}; the other relies on discrete data augmentations to simulate diverse data distributions \cite{UFR, Div}. However, domain shifts in the real world (e.g., weather or lighting changes) typically exhibit continuous and gradual characteristics \cite{domainlianxu, wu2025-OOD2}, such as transitions from sunny to cloudy to rainy weather. As a result, neither discrete data augmentations nor static textual prompts can capture the continuous evolution of cross-domain features \cite{wu4-OOD}, limiting the model's ability to perceive fine-grained inter-domain differences and, consequently, restricting generalization performance. Moreover, existing textual prompt methods depend on target domain textual priors \cite{SECOT}, which conflict with the core assumption of Single-DGOD, further limiting their practical applicability.

To address these challenges, we propose a novel method, i.e., Liquid Temporal Feature Evolution, which simulates the continuous feature evolution trajectory from the source domain to simulated latent distributions (top of Fig. \ref{F1}) by incorporating temporal modeling and liquid neural network-driven parameter adjustment. This method captures the progressive nature of cross-domain feature evolution, thereby improving generalization to unknown domains. Our method is motivated by two insights: 1) The continuity of domain shifts \cite{gs2, gs1} can be modeled through progressive Gaussian perturbations \cite{DGS} to simulate feature evolution; 2) The dynamic adaptation capability of liquid neural networks \cite{LNN1, LNN2} allows adjusting feature evolution paths based on temporal information, facilitating smooth transition between the source domain and simulated distributions.
Specifically, we design a collaborative mechanism of multi-scale Gaussian blur and noise to generate continuously evolving features in the feature space (bottom of Fig. \ref{F1}), simulating the initial feature evolution trajectory from the source domain to simulated distributions. We then model temporal correlations in feature evolution using spatiotemporal memory units, capturing long-range dependencies in the feature sequence. Finally, leveraging liquid neural networks and dynamic evolution information, we construct feature adjustment parameters to achieve smooth feature evolution between the source domain and simulated distributions, enhancing adaptability and robustness to domain shifts. Additionally, to avoid disrupting target features \cite{wu4-OOD}, we constrain the feature evolution process using intra-class consistency and inter-class separability losses, ensuring target features are preserved during evolution from the source domain to simulated distributions.

To evaluate the generalization capability of the proposed method under different types of domain shifts, we conducted experiments on both continuous (Diverse Weather dataset) and discontinuous (Real-to-Art benchmark) domain shift scenarios. Although our method is primarily designed for continuous domain shifts, it also achieves strong performance in discontinuous settings, demonstrating robust generalization. This can be attributed to the rich perturbation space constructed by multi-scale Gaussian blurring and controllable noise injection, as well as the dynamic adaptation of feature evolution paths enabled by the liquid neural network. These components allow the model to flexibly adapt to both gradual and abrupt distribution shifts, exhibiting strong robustness across diverse domain shift patterns.





\vspace{-5pt}
\section{Related Work}

\label{sec:formatting}
\subsection{Single-domain Generalization Object Detection}
Single-domain generalized object detection (Single-DGOD) is more challenging than multi-domain or domain-adaptive tasks, as it requires robust generalization to multiple unseen target domains using only a single source domain. While single-domain generalization (SDG) has advanced in image classification \cite{C1, C2, C3} and semantic segmentation \cite{Poda, sg2, sg3}, progress in object detection lags.
Existing Single-DGOD methods fall into two categories: Vision language approaches \cite{C_Cap, Poda} use style prompts to estimate domain changes but fail to capture continuous cross-domain feature variations, miss subtle differences, and contradict the premise of the ‘unknown target domain’. Static data augmentation methods (e.g., \cite{Div, G-NAS}) enhance diversity via image perturbations but cannot simulate deep semantic cross-domain variations.
To address these, we simulate the evolutionary trajectory of feature distributions from the source to latent domains, capturing continuous cross-domain variations to strengthen generalization to unknown domains.

\begin{figure*}[!t]
  \centering
  \includegraphics[width=2.0\columnwidth]{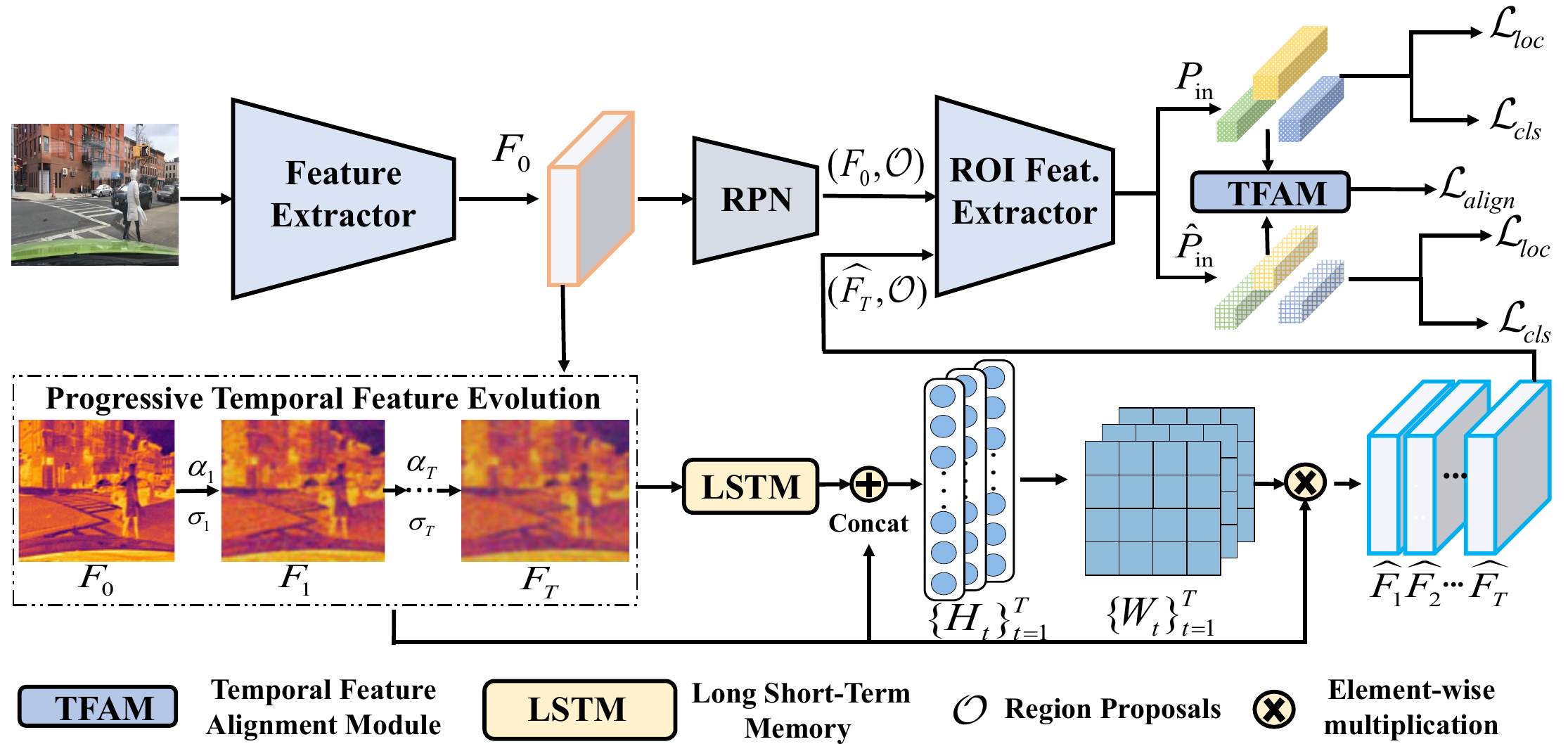}
  
  \caption{\textbf{Illustration of Liquid Temporal Feature Evolution.} First, the first-layer feature map $F_0$ extracted by the backbone network is iteratively perturbed to generate the initial feature sequence $\{F_t\}_{t=1}^T$, simulating the feature transitions from the source domain to a latent domain. Next, LSTM is employed to model the generated feature sequence, capturing the spatiotemporal dependencies of feature evolution. Then, liquid parameter evolution is applied to dynamically adjust the original feature sequence, yielding the final evolved sequence \(\{\hat{F}_t\}_{t=1}^T\), enabling a smooth and continuous transition between the source domain and the potential domain distributions. Finally, an alignment loss is used to further constrain the entire feature evolution process.} 
    \label{F2}
    \vspace{-8pt}
\end{figure*}

\vspace{-5pt}
\subsection{Liquid Neural Network}

Liquid Neural Networks \cite{LNN1} have gained attention for their ability to dynamically adapt to changing environments and process temporal information. Unlike traditional fixed-architecture networks, they simulate the continuous evolution of information through dynamic recurrent connections, allowing them to handle complex, time-varying data. Their key advantage is enhanced robustness and generalization when facing input noise, environmental changes, and data uncertainty. While primarily applied in robotic control and time-series forecasting \cite{LNN2, LNN3}, their use in object detection and domain generalization remains limited. In this paper, we leverage the dynamic adaptability and generalization of Liquid Neural Networks to dynamically adjust the evolving feature trajectory, enabling a smooth and continuous transition between the source domain and latent domain distribution.

\section{Method}
As shown in Figure \ref{F2}, to address the Single-Domain Generalized Object Detection (Single-DGOD) problem, we propose a Liquid Temporal Feature Evolution (LTFE) method. By simulating the evolutionary trajectory of feature distributions from the source domain to the simulated latent distributions, it captures the continuous variation patterns of features across domains, thereby enhancing the model's generalization ability in the unknown domains. The overall training and inference process is shown in Algorithm 1.

\subsection{Progressive Temporal Feature Evolution}
To simulate the progressive evolution of features from a source domain to a latent distribution, we introduce a sequential perturbation mechanism that models the underlying feature shift. We follow the baseline work and exploit the widely used object detector, i.e., Faster R-CNN \cite{ren2015faster}, as the basic detection model. Given an input image, the backbone network (e.g. ResNet) extracts an initial feature map \( F_0 \in \mathbb{R}^{w \times h \times c} \), where \( w \), \( h \), and \( c \) denote the width, height, and channel dimension, respectively. Experimental results show that evolving features in the first layer and using them to generate layers 2–4 leads to the best performance, as detailed in the supplementary material.

To emulate cross-domain transition, we progressively perturb the features. At each time step \( t \), the feature map is transformed by applying a combination of Gaussian blurring \cite{DGS} and stochastic noise injection:
\begin{eqnarray}
F_t = G(\sigma_t) * F_{t-1} + \alpha_t \cdot G(\sigma_t),
\end{eqnarray}
where \( \sigma_t \) controls the degree of blurring, \( \alpha_t \) determines the noise intensity, and \( G(\sigma_t) \) denotes the Gaussian blur kernel with variance \( \sigma_t \), defined as:
\begin{eqnarray}
G(\sigma_t) = \frac{1}{2\pi\sigma_t^2} \exp\left(-\frac{i^2 + j^2}{2\sigma_t^2}\right),
\end{eqnarray}
where \( (i, j) \) represents the spatial coordinates in the kernel. The first term progressively blurs the image to simulate visual degradation, while the second injects Gaussian noise with exponentially decaying intensity:
\begin{eqnarray}
\alpha_t = \alpha_0 \cdot \exp(-\lambda t),
\end{eqnarray}
where \( \alpha_t \) follows an exponential decay to ensure that noise influence diminishes over time, aligning with realistic domain shift patterns. And \( \lambda \) is a hyperparameter set to 0.2, controlling the rate at which the noise intensity decays over time. The baseline noise intensity $\alpha_0$ is set to 0.2, while the blurring intensity increases exponentially:  
\begin{eqnarray}
\sigma_t = \sigma_0 \cdot \gamma^t,
\end{eqnarray}
where \( \gamma \) is a hyperparameter set to 1.2, controlling the degradation rate, ensuring that features transition progressively from a fine-grained state (\( \sigma_0 = 1 \)) to a coarser representation (\( \sigma_t \) increasing).

\subsection{Temporal Dependency Modeling}
To capture temporal dependency patterns in feature evolution trajectories, we feed the simulated feature sequence $\{F_t\}_{t=1}^T$ generated by temporal perturbations into an LSTM network \cite{LSTM} for spatiotemporal interaction modeling. The memory state propagation is implemented through the following gated mechanism:  
\begin{eqnarray}
h_t, c_t &= \text{LSTM}(F_t, h_{t-1}, c_{t-1}),
\end{eqnarray}
where $h_t \in \mathbb{R}^d$ and $c_t \in \mathbb{R}^d$ denote the hidden state and cell state at timestep $t$, respectively. The hidden state $h_t$ captures feature abstractions at the current time step, while the cell state $c_t$ maintains long-term memory. To enhance the interaction between original features and historical memory, we introduce a feature-state fusion mechanism:
\begin{eqnarray}
H_t &= \text{ReLU}(W_p [h_t \oplus F_t]),
\end{eqnarray}
where $W_p \in \mathbb{R}^{(d+c)\times d}$ represents a learnable projection matrix, and $\oplus$ indicates channel-wise concatenation.  The final temporal encoding features $H = \{H_t\}_{t=1}^T \in \mathbb{R}^{T \times d}$, where $d$ is the hidden dimension, comprehensively capture the evolution patterns from the initial state $F_1$ to the final state $F_T$. This feature sequence offers spatio-temporally consistent representations for generating dynamic parameters.

\begin{algorithm}[t]
   \caption{Liquid Temporal Feature Evolution}
   \KwIn{Input image \(I\), initial feature map \(F_0\), hyperparameters \(\alpha_0, \lambda, \gamma\), number of steps \(T\)}
   \KwOut{Evolved feature sequence \(\{\hat{F}_t\}_{t=1}^T\)}

   \textbf{Training Phase:}
   \While{train}{
       Initialize \(F_0\)\;
       \For{$t = 1$ \KwTo $T$}{
           \(\sigma_t = \sigma_0 \cdot \gamma^t\), \(\alpha_t = \alpha_0 \cdot \exp(-\lambda t)\)\;
           \(F_t = G(\sigma_t) * F_{t-1} + \alpha_t \cdot G(\sigma_t)\)\;
       }
       Feed \(\{F_t\}_{t=1}^T\) into LSTM and obtain \(\{H_t\}_{t=1}^T\)\;
       \For{$t = 1$ \KwTo $T$}{
           \(H_t = \text{ReLU}(W_p [h_t \oplus F_t])\)\;
       }
       Solve ODE: \(W_t = \text{ODESolve}(f_\theta, W_0, H_t, t)\)\;
       Adjust the feature: \(\hat{F}_t = \text{Conv2D}(F_0, W_t) + F_0\)\;
}

   \textbf{Inference Phase:}
   \While{eval}{
       Extract feature \(F_0\) from input image \(I\)\;
       Simulate temporal feature perturbations for 1-2 steps to generate \(\{F_t\}_{t=1}^2\)\;
       Generate dynamic convolution kernel \(W_{\text{test}} = \text{ODESolve}(f_\theta, W_0, H_t,  \tau)\)\;
       Adjust feature map: \(\hat{F}_t = \text{Conv2D}(F_0, W_{\text{test}}) + F_0\)\;
      Classify and localize using \(\hat{F}_T\)\.;
   }
\end{algorithm}

\subsection{Liquid Parameter Evolution}
To ensure a continuous and smooth transition between the source and latent distribution, we construct a dynamic convolution kernel generation network based on Neural Ordinary Differential Equations \cite{NODA} to adjust the evolved features, as described in the following equation:
\begin{eqnarray}
\frac{dW(\tau)}{d\tau} = f_\theta \left( W(\tau), H_t \right),
\end{eqnarray}
where $f_\theta: \mathbb{R}^d \rightarrow \mathbb{R}^{k \times k \times c_{\text{in}} \times c_{\text{out}}}$ is a vector field function implemented using a liquid neural network, $H_t \in \mathbb{R}^d$ denotes the temporal feature encoding at time step $t$, and $W(\tau)$ is the convolution kernel parameter evolving along a continuous trajectory governed by virtual time $\tau$. The kernel size is $k \times k$, and $c_{\text{in}}$, $c_{\text{out}}$ are the input and output channels.

The fourth-order Runge-Kutta \cite{RK} method is used to numerically solve the ODE \cite{NODA}, generating the time-varying convolution kernel. The process is described as follows:
\begin{eqnarray}
W_t = \text{ODESolve}\left( f_\theta, W_0, H_t, \tau \right),
\end{eqnarray}
where $W_0 \in \mathbb{R}^{k \times k \times c_{\text{in}} \times c_{\text{out}}}$ is the initial kernel (inherited from the pre-trained Faster R-CNN), and the integration interval $\tau \in [0, \hat{\tau}]$ is dynamically determined by the L2 norm of the current temporal feature encoding $H_t$:
\begin{eqnarray}
\hat{\tau} = \frac{\|H_t\|_2}{\max\|H_t\|_2},
\end{eqnarray}
which normalizes evolution time by the sequence’s maximum feature magnitude, ensuring larger temporal deviations (higher $\|H_t\|_2$) result in longer kernel evolution and improved adaptation to significant domain shifts. The dynamically generated kernel is then applied to the initial feature $F_0$ for input-conditioned adjustment:
\begin{eqnarray}
\hat{F_t} = \text{Conv2D}\left( F_0, W_t \right) + F_0,
\end{eqnarray}
where the residual connection preserves original semantic information. With a fixed $k \times k$ ($k=3$) convolution kernel $W_t$ and consistent channels, a dynamic feature sequence $\{\hat{F}_t\}_{t=1}^T$ is produced, providing continuous training signals. Though only $\hat{F}_T$ is used in inference, LSTM-based temporal modeling of intermediate features enables learning of domain shift dynamics, ensuring $\hat{F}_T$ reflects the full evolution trajectory and improves generalization to unseen domains.

\subsection{Temporal Feature Alignment Module}
To ensure semantic consistency of target features during feature evolution and mitigate noise interference, we introduce feature alignment using intra-class consistency and inter-class separability losses. Specifically, the original feature $F_0$ is fed into the RPN to generate proposals $\mathcal{O}$. Then, both $F_0$ and the final evolved feature $\hat{F}_T$, along with $\mathcal{O}$, are passed through the ROI head to extract proposal-level features $P_{\text{in}} \in \mathbb{R}^{m \times n}$ and $\hat{P}_{\text{in}} \in \mathbb{R}^{m \times n}$, where $m$ and $n$ denote the number of proposals and feature dimensions. Notably, the intermediate feature sequence models continuous domain shifts via progressive perturbation and temporal dynamics, while the alignment loss enforces consistency in target semantics throughout the evolution.

Next, we minimize the feature distance of same-class instances across domain evolution using \( P_{\text{in}} \) and \( \hat{P}_{\text{in}} \):
\begin{eqnarray}
\mathcal{L}_{\text{intra}} = \frac{1}{N} \sum_{i=1}^N \| P_{\text{in}}^{(i)} - \hat{P}_{\text{in}}^{(i)} \|_2^2.
\end{eqnarray}
At the same time, we maximize the feature differences between instances of different classes as follows:
\begin{eqnarray}
\mathcal{L}_{\text{inter}} = -\log \frac{\exp(s(P_{\text{in}}^{(i)}, \hat{P}_{\text{in}}^{(i)}))}{\sum_{j \neq i} \exp(s(P_{\text{in}}^{(i)}, \hat{P}_{\text{in}}^{(j)}))}, 
\end{eqnarray}
where \( s(\cdot) \) is the cosine similarity function. The total alignment loss is defined as:
\begin{eqnarray}
\mathcal{L}_{\text{align}} = \lambda_1 \mathcal{L}_{\text{intra}} + \lambda_2 \mathcal{L}_{\text{inter}},
\end{eqnarray}
which \(\lambda_1\) and \(\lambda_2\) are weight balancing coefficients, with \(\lambda_1 = 1.0\) and \(\lambda_2 = 0.1\) as fixed values. Then, we input the example features \( P_{\text{in}} \) and \( \hat{P}_{\text{in}} \) into the object classifier and regressor to calculate the classification loss \( \mathcal{L}_{\text{cls}} \) and localization loss \( \mathcal{L}_{\text{loc}} \). The joint objective is defined as:
\begin{eqnarray}
\mathcal{L}_{\text{total}} = \mathcal{L}_{\text{cls}} + \mathcal{L}_{\text{reg}} + \mathcal{L}_{\text{align}},
\end{eqnarray}
where \( \mathcal{L}_{\text{cls}} \) and \( \mathcal{L}_{\text{reg}} \) are the classification and regression losses, respectively.

\subsection{Inference for Target Domain Object Detection}
During inference, unlike training, the model requires only minimal feature evolution, performing limited perturbations to fine-tune adaptation to the real target domain and bridge the gap between simulated training and true distributions. Typically, two discrete iterations ($T=2$) suffice for adaptive feature adjustment. The process begins with image feature extraction to generate the initial evolved feature sequence $\{F_t\}_{t=1}^2$, which is then refined via the Temporal Dependency Modeling and Liquid Parameter Evolution. In this mechanism, the Liquid Neural Network (LNN) employs ODE-based parameter evolution to dynamically generate convolution kernels, not as random augmentation, but as a controlled adjustment to maintain the semantic consistency of target features during perturbation. The final adjusted feature $\hat{F}_2$ is used for classification and localization. This approach enables targeted feature evolution and dynamic adaptation during inference, allowing the model to align with the target domain’s feature distribution while preserving critical semantic information. Algorithm 1 details the training and inference procedures. 
 \vspace{-5pt}
\section{Experiments}

In the experiments, for Single-DGOD, we follow the settings of the work \cite{S_DG, pdoc, UFR} to validate the model's generalization capability. Furthermore, to validate the effectiveness of our method, we evaluated the model on the Reality-to-Art generalization benchmark \cite{Div}. 
\vspace{-5pt}
\subsection{Experimental setup}

\textbf{Dataset. }The entire experiment is conducted based on the Diverse-Weather and Real-to-Art benchmarks. 
\textbf{Diverse Driving Weather Scenarios.} 
To ensure a fair comparison, we use the same training and testing datasets as other Single-DGOD methods \cite{S_DG, UFR, Div}. Training is conducted only on the daytime sunny dataset, followed by direct testing on four target domains with different weather conditions: nighttime clear, dusk rainy, nighttime rainy, and daytime foggy. The category space remains consistent across all datasets.
\textbf{Generalization from Reality to Art.}  
Following \cite{Art, Div}, we train on the Pascal VOC2007 and VOC2012 trainval sets \cite{VOC}, and evaluate generalization on Clipart1k, Watercolor2k, and Comic2k. Clipart shares all 20 Pascal VOC classes, while Watercolor2k and Comic2k include 6-class subsets.

\textbf{Metric.}  
We adopt the same evaluation metrics as Single-DGOD \cite{S_DG, UFR} to ensure a fair comparison with other methods. We use Mean Average Precision (mAP) with an IoU threshold of 0.5 to evaluate the model's performance across various datasets and categories. Specifically, we compute mAP@0.5 for all categories in each test dataset to evaluate the model's generalization capability across diverse scenes and object classes.
\vspace{-15pt}
\subsection{Implementation Details}  
\vspace{-2pt}
For a fair comparison with other Single-DGOD methods \cite{S_DG, UFR}, we use the same baseline model, Faster R-CNN \cite{ren2015faster}. While prior works report results only with ResNet-101 \cite{resnet}, we provide a more comprehensive evaluation using ResNet-50, ResNet-101, and Swin Transformer \cite{swin}. The model achieves optimal performance during training when \( T = 8 \) in the Progressive Temporal Feature Evolution process. In Equation (1), \( \alpha_0 \) and \( \sigma_0 \) are initialized to 0.2 and 1, respectively. For inference, \( T \) is set to 2 to preserve target domain information. Training is conducted with SGD (momentum = 0.9) for 20 epochs, starting with a learning rate of 0.02. All experiments run on two NVIDIA RTX 3090 GPUs (24GB) with images resized to 800×800 pixels.
\vspace{-2pt}

\begin{table}[!t]
  \centering

  \resizebox{\linewidth}{!}{
    \begin{tabular}{cc|c|cccc}
    \toprule
    \toprule
    \multicolumn{2}{c|}{\multirow{2}[4]{*}{Method}} & \multicolumn{1}{c}{} &       & \multicolumn{1}{c}{mAP} &       &  \\
\cmidrule{3-7}    \multicolumn{2}{c|}{} & \multicolumn{1}{c|}{Day Clear} & \multicolumn{1}{c}{Night Sunny} & \multicolumn{1}{c}{Dusk Rainy} & \multicolumn{1}{c}{Night Rainy} & \multicolumn{1}{c}{Day Foggy} \\
    \midrule
    \multicolumn{2}{c|}{Faster R-CNN(Res101) \cite{ren2015faster}} & 48.1  & 34.4  & 26.0  & 12.4  & 32.0  \\
    \multicolumn{2}{c|}{SW(Res101) \cite{SW}} & 50.6  & 33.4  & 26.3  & 13.7  & 30.8  \\
    \multicolumn{2}{c|}{IBN-Net(Res101) \cite{IBN}} & 49.7  & 32.1  & 26.1  & 14.3  & 29.6  \\
    \multicolumn{2}{c|}{IterNorm(Res101) \cite{Interfrom}} & 43.9  & 29.6  & 22.8  & 12.6  & 28.4  \\
    \multicolumn{2}{c|}{ISW(Res101) \cite{ISW}} & 51.3  & 33.2  & 25.9  & 14.1  & 31.8  \\
    \multicolumn{2}{c|}{S-DGOD (Res101)\cite{S_DG}} & 56.1 & 36.6  & 28.2  & 16.6  & 33.5  \\
    \multicolumn{2}{c|}{C-Gap (Res101)\cite{C_Cap}} & 51.3  & 36.9  & 32.3  & 18.7  & 38.5  \\
    \multicolumn{2}{c|}{PDOC (Res101)\cite{pdoc}} & 53.6  & 38.5  & 33.7  & 19.2  & 39.1  \\
    \multicolumn{2}{c|}{UFR (Res101)\cite{UFR}} & 58.6  & 40.8  & 33.2  & 19.2  & 39.6  \\
    \multicolumn{2}{c|}{DIV (Res101) \cite{Div}} & 52.8  & 42.5  & 38.1  & 24.1  & 37.2  \\
    \multicolumn{2}{c|}{G-NAS (Res101) \cite{G-NAS}} & 58.4  & 45.0  & 35.1  & 17.4  & 36.4  \\
     \multicolumn{2}{c|}{SECOT (Res101) \cite{SECOT}} &  55.4  & 42.0  & 39.2  & 24.5  &  40.6  \\
    \multicolumn{2}{c|}{FWCL (Res101) \cite{FWCL}} &  55.5  & 37.5  & 32.6  & 18.9  &  32.3  \\
    \midrule
    \multicolumn{2}{c|}{Ours(Res50)} & 50.9 & 31.2  & 29.4  & 16.7  & 35.4  \\
   
     \multicolumn{2}{c|}{Ours(Res101)} &  58.4  & 43.1  & 39.7  & 24.3  & 41.2\\
    \multicolumn{2}{c|}{Ours(Swin-T)} & \textbf{64.2} & \textbf{53.1} & \textbf{46.5} & \textbf{33.4} & \textbf{46.4} \\
    \bottomrule
    \bottomrule
    \end{tabular}%
    
    }
    \vspace{-0.05in}
  \caption{Single-Domain Generalization Results (mAP(\%)) in Diverse Weather Driving Scenarios. \textbf{The bold sections} represent the best results.}
    \vspace{-15pt}
  \label{tab1}%
\end{table}%
\begin{figure*}[!t]
  \centering
  \includegraphics[width=2.0\columnwidth]{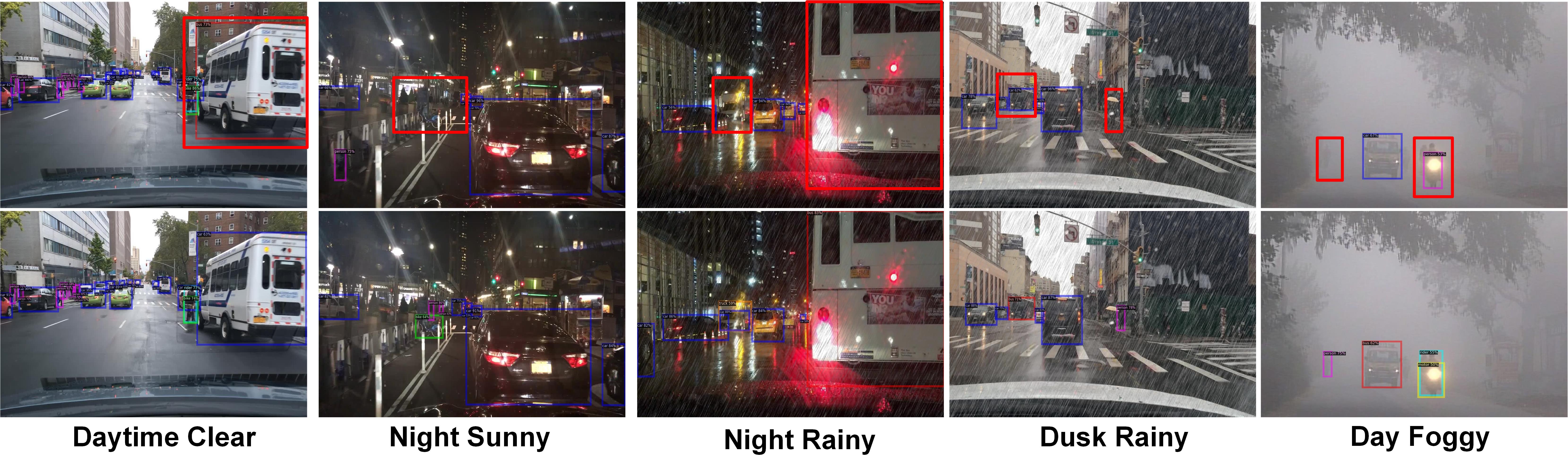}
  \vspace{-10pt}
  \caption{\textbf{Qualitative Results:} Detection results under different weather conditions. The first and second rows display the results from G-NAS \cite{G-NAS} and our method, respectively. To provide a more intuitive comparison, we highlight the objects missed or incorrectly detected by G-NAS \cite{G-NAS} using red boxes, which are correctly identified by our method.}
  \vspace{-10pt}
    \label{f5}
\end{figure*}

\vspace{-3pt}
\subsection{Comparison with the State of the Art}
\textbf{Diverse Driving Weather Scenarios.} We compare our method with state-of-the-art Single-DGOD methods \cite{UFR, G-NAS}, as shown in Table \ref{tab1}. Under scenarios with continuous domain bias, our method achieves superior performance across all four target domains. Figure \ref{f5} presents detection results under four weather conditions, demonstrating that our model maintains strong performance even in challenging scenarios such as rainy nights. Detailed per-class performance and metric analysis for all target domains are provided in the supplementary material.

\begin{table}[h]
  \centering
   
  \resizebox{\linewidth}{!}{
    \begin{tabular}{cc|c|ccc}
    \toprule
    \toprule
    \multicolumn{2}{c|}{\multirow{2}[4]{*}{Method}} & \multicolumn{1}{c}{} &       & \multicolumn{1}{c}{mAP} &  \\
\cmidrule{3-6}    \multicolumn{2}{c|}{} & \multicolumn{1}{c|}{VOC} & \multicolumn{1}{c}{Comic} & \multicolumn{1}{c}{Watercolor} & \multicolumn{1}{c}{Clipart} \\
    \midrule
    \multicolumn{2}{c|}{Faster R-CNN(Res101)\cite{ren2015faster}} & 80.4  & 19.4  & 45.6  & 26.5  \\
    \multicolumn{2}{c|}{NP(Res101)\cite{np}} & 79.2  & 28.9  & 53.3  & 35.4  \\
    \multicolumn{2}{c|}{C-Gap(Res101)\cite{C_Cap}} & 80.5  & 29.4  & 50.7  & 36.7  \\
    \multicolumn{2}{c|}{DIV(Res101)\cite{Div}} & 80.1  & 33.2  & 57.4  & 38.9  \\
    \multicolumn{2}{c|}{SECOT(Res101)\cite{SECOT}} & 82.9  & 34.8  & 57.5  & 40.2  \\
    \midrule
    \multicolumn{2}{c|}{Ours(Res50)} & 80.2  & 27.4  & 54.1  & 35.6  \\
    \multicolumn{2}{c|}{Ours(Res101)} & 84.6  & 35.4  & 58.2  & 42.1  \\
    \multicolumn{2}{c|}{Ours(Swin-T)} & \textbf{88.2} & \textbf{37.3} & \textbf{61.3} &\textbf{44.5}\\
    \bottomrule
    \bottomrule
    \end{tabular}%
    }
  \label{t2}%
  \vspace{-0.09in}
  \caption{Single-Domain Generalization Results (mAP(\%)) from Real to Artistic. \textbf{Bold values} indicate the best results.}
  \label{t2}
  \vspace{-7pt}
\end{table}%

\textbf{Generalization from Reality to Art.}
We evaluate the generalization ability of our method across real and artistic domains, a representative case of non-continuous domain bias with large distribution gaps. As shown in Table \ref{t2}, using the same ResNet-101 backbone, our model consistently outperforms others on three artistic-style datasets. For instance, compared to the SECOT method \cite{SECOT}, it achieves a 4.7\% improvement on the Cliparts dataset. These results highlight the strong generalization and robustness of our approach under non-continuous domain shifts. Detailed per-class metrics and performance analysis are provided in the supplementary material.

\textbf{Comparison with Continual Test-Time Adaptation Methods.}
Since we introduced parameter adjustment during feature evolution to control the trajectory from the source domain to the target domain, we can adapt to the target domain distribution by fine-tuning parameters during the test phase. Therefore, we compared our method with the latest continual test-time adaptation methods \cite{SKIP, lideng}, as shown in Table \ref{TTA}. Compared to the test-time adaptation method, our method achieved the best performance in the four target domains, with a notable improvement in the night-sunny domain (from 35.3\% to 43.1\%) with ECPG \cite{lideng}. Furthermore, compared to direct testing without adaptation (Direct-Test in Table \ref{TTA}), our method demonstrates that feature evolution during training effectively enhances target domain adaptation and significantly improves generalization.

\begin{table}[!t]
  \centering

   \fontsize{20}{20}\selectfont
  \resizebox{\linewidth}{!}{
    \begin{tabular}{cc|ccccc}
    \toprule
    \toprule
    \multicolumn{2}{c|}{\multirow{2}[4]{*}{Method}} & \multicolumn{1}{c}{} &       & \multicolumn{1}{c}{mAP} &       &  \\
\cmidrule{3-7}    \multicolumn{2}{c|}{} 
& \multicolumn{1}{c}{Night Sunny} & \multicolumn{1}{c}{Dusk Rainy} & \multicolumn{1}{c}{Night Rainy} & \multicolumn{1}{c}{Day Foggy} \\
    \midrule
   \multicolumn{2}{c|}{Baseline}   & 35.4  & 27.5  & 13.7  & 32.6  \\
   \multicolumn{2}{c|}{MemCLR \cite{mem}}   & 33.5  & 32.7  & 16.4  & 36.7  \\
    \multicolumn{2}{c|}{SKIP \cite{SKIP}}   & 32.6  & 34.1  & 17.3  & 32.9  \\
   \multicolumn{2}{c|}{ECPG \cite{lideng}}   & 35.3  & 35.7  & 18.2  & 33.7  \\
    \midrule
    \multicolumn{2}{c|}{Direct-Test}  & 42.4  & 37.9  & 23.6  & 38.9\\
     \multicolumn{2}{c|}{Ours}   & \textbf{43.1}  & \textbf{39.7}  & \textbf{24.3}  & \textbf{41.2}\\
    \bottomrule
    \bottomrule
    \end{tabular}%
    
    }
    \caption{Continual test-time adaptive object detection results (\%) on the Diverse-Weather dataset using the ResNet-101.}
    \vspace{-15pt}
  \label{TTA}%
\end{table}%

\begin{table}[!h]
\centering
 \vspace{-5pt}
\resizebox{1.0\linewidth}{!}{
\begin{tabular}{lcccc}
\toprule
\toprule
\textbf{Model} & \textbf{mAP (\%)} & \textbf{Params (M)} & \textbf{FPS} & \textbf{FLOPs (G)} \\
\hline
Faster R-CNN \cite{ren2015faster} & 26.2   & 90.8   & 4.7   & 267 \\
S-DGOD \cite{S_DG}               & 28.7   &   120.4  & 3.9  &  284  \\
C-Gap \cite{C_Cap}               & 31.6   & 190.2    & 2.5  & 423  \\
PDOC \cite{pdoc}                 & 32.6   & 198.6   & 2.2   & 478 \\
UFR \cite{UFR}                   & 33.2   & 184.2   & 6.9   & 323 \\
 \midrule
Ours(Res101)       & 37.1   & 140.6    & 6.4  & 325  \\
\bottomrule
    \bottomrule

\end{tabular}
}
 \vspace{-5pt}
\caption{Model Comparison in Terms of mAP, Parameters, FPS, and FLOPs. The mAP refers to the average mAP (\%) across the four target domains.}
 \vspace{-5pt}
\label{FPS}%
\end{table}

\textbf{Computational Efficiency and Time Complexity Analysis.}  
We compare the time complexity and computational efficiency of existing methods in Table \ref{FPS}. As can be seen, compared to other single-domain generalization methods, such as C-CAP\cite{C_Cap} and PDOC\cite{pdoc}, our approach achieves superior performance with lower computational cost and reduced latency, demonstrating its efficiency and deployability.

\begin{table}[!t]
  \centering

  \fontsize{22}{22}\selectfont
     \resizebox{\linewidth}{!}{
    \begin{tabular}{ccc|c|cccc}
    \toprule
    \toprule
    \multicolumn{3}{c|}{Method} & \multicolumn{1}{c|}{Source} & \multicolumn{4}{c}{Target} \\
    \midrule
    \multicolumn{1}{c}{PTFE} & \multicolumn{1}{c}{TDM+LPE}  & \multicolumn{1}{c|}{TFAM} & \multicolumn{1}{c|}{Day Clear} &  \multicolumn{1}{c}{Night Sunny}& \multicolumn{1}{c}{Dusk Rainy} & \multicolumn{1}{c}{Night Rainy}  & \multicolumn{1}{c}{Day Foggy}  \\
    
    \midrule
         &        &       &  49.2  & 35.4  & 27.5  & 13.7  & 32.6\\
    $\checkmark$     &           &       & 54.6  & 36.8  & 29.3  & 14.1  & 33.6 \\
    $\checkmark$     &  $\checkmark$         &       & 57.3  & 42.6  & 39.4  & 22.7  & 40.8  \\  
    
    $\checkmark$     &      &  $\checkmark$     & 56.8  & 40.2  & 34.1  & 19.7  & 36.5 \\
    $\checkmark$     & $\checkmark$       & $\checkmark$     & \textbf{58.4} & \textbf{43.1} & \textbf{39.7} & \textbf{24.3} & \textbf{41.2} \\
    \bottomrule
    \bottomrule
    \end{tabular}%
    }
  
   \caption{\textbf{Ablation analysis of our proposed model.} PTFE stands for  Progressive Temporal Feature Evolution, TDM + LPE denotes the combination of Temporal Dependency Modeling and Liquid Parameter Evolution, and TFAM denotes Temporal Feature Alignment Module.}
   \vspace{-15pt}
   \label{teb6}%
\end{table}%

\begin{figure*}[!h]
  \centering
  \includegraphics[width=1.95\columnwidth]{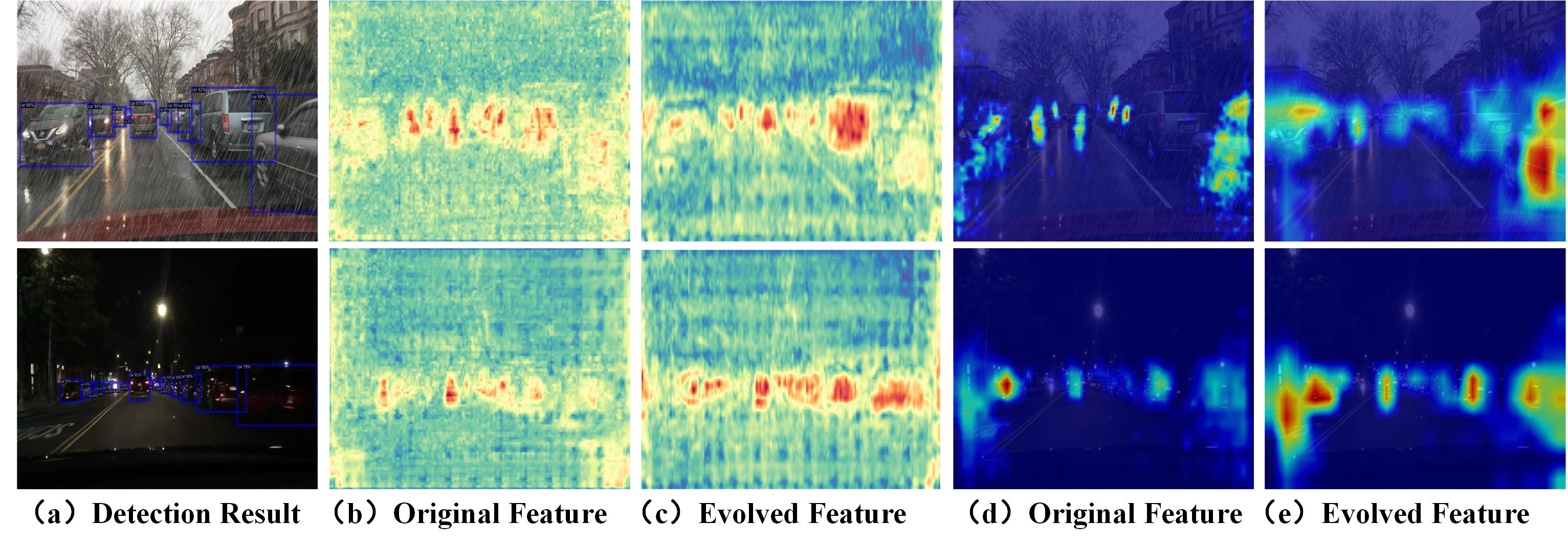}
  \vspace{-8pt}
  \caption{\textbf{Visualization analysis of our method}: The first column displays the model's detection results. The second and third columns correspond to the channel activation visualizations of the initial feature \( F_0 \) and the evolved feature \( \hat{F}_T \), respectively. The fourth and fifth columns display the heatmap visualizations of the initial feature \( F_0 \) and the evolved feature \( \hat{F}_T \). }
    \label{vis}
    \vspace{-14pt}
\end{figure*}

\subsection{Ablation Study} 
\textbf{Component Analysis.} To validate the effectiveness of each component, we conducted ablation studies on Progressive Temporal Feature Evolution (PTFE), the combined Temporal Dependency Modeling (TDM) and Liquid Parameter Evolution (LPE) module, and Temporal Feature Alignment Module (TFAM). Since LPE relies on temporal information for dynamic adjustment, it is evaluated jointly with TDM. Results are shown in Table \ref{teb6}, with the first row representing the Faster R-CNN baseline.
The second row adds PTFE to simulate the initial feature evolution without dynamic optimization. The third row further incorporates TDM and LPE, effectively capturing cross-domain variations and increasing data diversity during training, leading to significant performance gains over the second row. The fourth row adds TFAM to PTFE to reduce target information loss during evolution. The final row shows the full model, where dynamic control of the feature evolution trajectory better approximates the distribution shift from source to potential target domains, significantly enhancing generalization. Detailed ablation analyses are provided in the supplementary material.

\begin{figure}[!t]
  \centering
  \includegraphics[width=1.0\columnwidth]{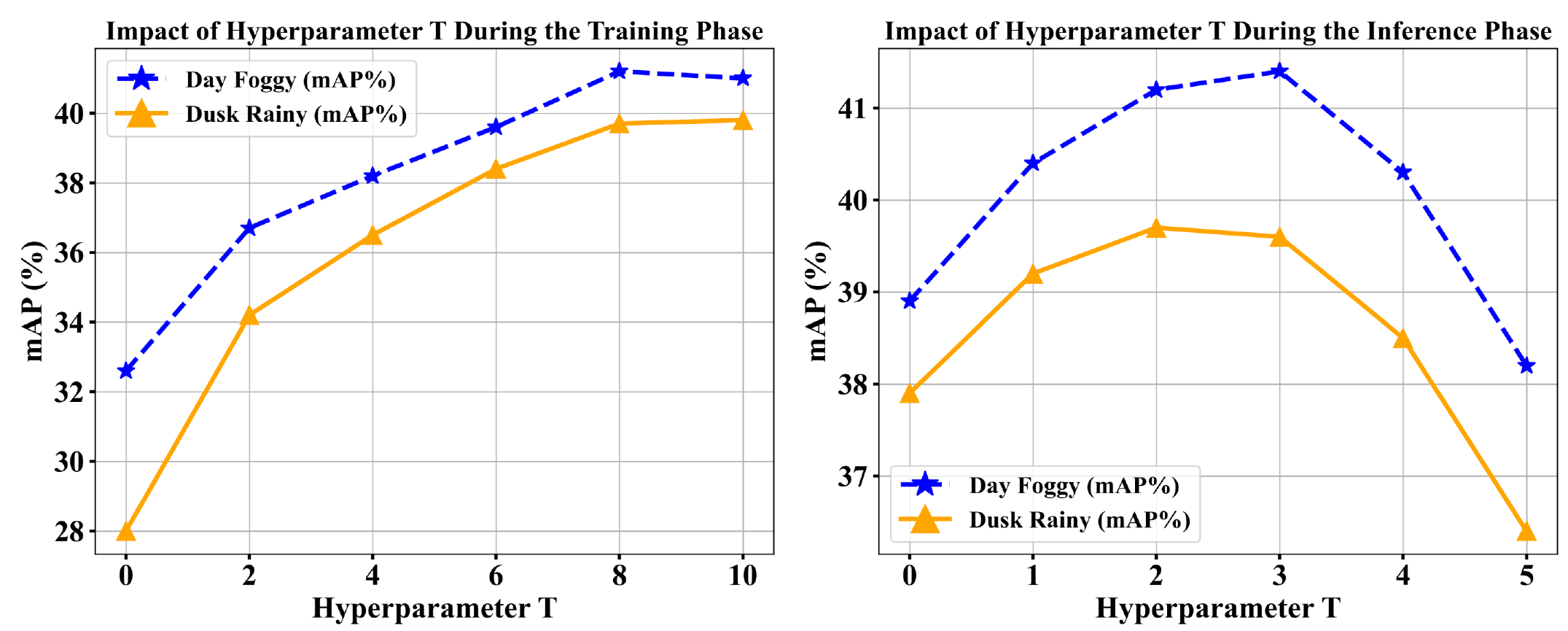}
   \vspace{-15pt}
  \caption{Analysis of the time step \( T \) in the Progressive Temporal Feature Evolution.}
    \label{T}
       \vspace{-16.5pt}
\end{figure}

\textbf{The Impact of Time Step \( T \) in the Evolution of Progressive Temporal Feature Evolution.} The time step \(T\) is a crucial hyperparameter in Progressive Temporal Feature Evolution (PTFE) that controls both the duration and granularity of the feature evolution process. During training, increasing \(T\) extends the evolution trajectory, allowing for a more comprehensive simulation of domain shifts from the source to the target domain and enhancing the model's adaptability in cross-domain scenarios.
As shown in Figure \ref{T}, increasing the number of iterations \(T\) continuously improves the mAP(\%) on the target domains ‘Day Foggy’ and ‘Dusk Rainy,’ converging at around 8 iterations. However, during inference, an excessively large \(T\) may cause the evolved target domain features to deviate from the image's original semantic content while accumulating noise, ultimately degrading detection accuracy. As shown on the right of Figure \ref{T}, the optimal performance during inference is achieved with \(T\) set to approximately 2 iterations.

\begin{figure}[!t]
  \centering
  \includegraphics[width=0.92\columnwidth]{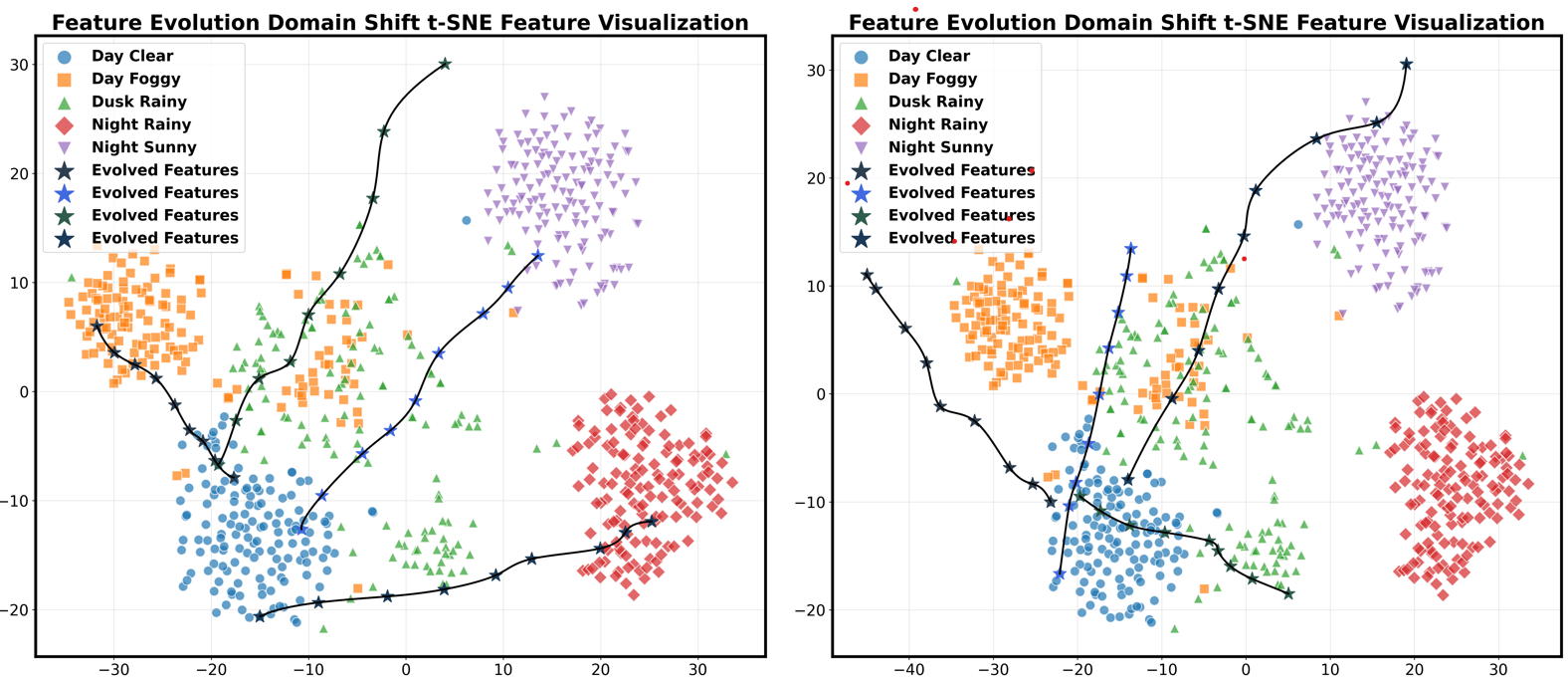}
 \vspace{-5pt}
  \caption{t-SNE Visualization Analysis of Liquid Temporal Feature Evolution.}
    \label{tsne}
     \vspace{-15pt}
\end{figure}

\textbf{t-SNE Visualization of Liquid Temporal Feature Evolution.} To validate the effectiveness of the progressively shifted feature evolution in our proposed Liquid Temporal Feature Evolution (LTFE), we conducted t-SNE visualization on the generated evolving feature sequence, as shown in Fig. \ref{tsne}. By mapping features from different time steps into a two-dimensional space, we can visually observe the evolution of features between the source and target domains. The visualization results show that as the time steps increase, features from the source domain progressively shift toward those of the target domain. This confirms that our method effectively simulates the data distribution of the potential target domain through gradual perturbations, validating the effectiveness of the proposed LTFE.

\vspace{-6pt}
\subsection{Visualization Analysis}

In Figure \ref{vis}, we present channel activation and heatmap visualizations to compare the initial feature \( F_0 \) with the final evolved feature \( \hat{F}_T \) during the testing phase. The results show that the final evolved feature \( \hat{F}_T \) exhibits stronger activation values across multiple channels, indicating that the model captures key features of the target domain more effectively during feature evolution. This enhanced activation highlights the model's attention to crucial information in the target domain, particularly in complex backgrounds or across different object categories. Furthermore, heatmap visualization shows that the final evolved feature $\hat{F}_T$ focuses more on foreground objects, indicating improved localization and recognition in detection tasks. These results confirm that simulating distribution shifts toward the latent data distribution enables progressive feature refinement, enhancing generalization to unseen domains.

\vspace{-5pt}
\section{Conclusion}

In the Single-DGOD task, we propose a novel method, Liquid Temporal Feature Evolution, which enhances the model's generalization capability in potential target domains by simulating the feature evolution trajectory from the source domain to unseen target domains. Our method leverages Progressive Temporal Feature Evolution, Temporal Dependency Modeling, and Liquid Parameter Evolution to achieve a smooth and continuous transition from the source domain to simulated
latent distributions. Additionally, we introduce intra-class consistency and inter-class separability losses to regulate the feature evolution process, ensuring that target features remain intact during the evolution process. Significant gains on the Diverse Weather dataset and Real-to-Art benchmark confirm its effectiveness.


\section{Acknowledgment}
This work is supported by the National Nature Science Foundation of China (Nos. 62376186, 62472333).

\bibliography{aaai2026}
\newpage
\appendix

\section{Additional Results}
\subsection{Further Details of the Dataset.}
To validate the effectiveness of the proposed method under different types of domain shift scenarios, we design experiments around two benchmark tasks, each corresponding to a distinct domain bias pattern: continuous domain shift (diverse driving weather scenarios) and non-continuous domain shift (generalization from reality to artistic styles). The experimental setups are detailed as follows:

\textbf{Diverse Driving Weather Scenarios (Continuous Domain Shift)}:
To ensure a fair comparison with existing studies, we adopt the same training and testing datasets used by other Single-Domain Generalized Object Detection (Single-DGOD) methods, such as SECOT \cite{SECOT}, UFR  \cite{UFR}, and DIV  \cite{Div}. The dataset consists of five subsets, each representing a distinct weather condition: daytime sunny, nighttime clear, dusk rainy, nighttime rainy, and daytime foggy.
In our setup, the model is trained solely on the daytime sunny subset, which features clear lighting conditions and a variety of traffic-related objects (e.g., vehicles, pedestrians, cyclists). This subset provides a stable and consistent source domain distribution. Once trained, the model is directly tested on the remaining four target domains, which simulate typical continuous domain shifts caused by illumination changes (nighttime clear), precipitation (dusk and nighttime rainy), and reduced visibility (daytime foggy).
It is important to note that all subsets share an identical category space, ensuring that any performance variation stems exclusively from domain shifts rather than class differences. This enables a precise evaluation of the method’s adaptability to weather-related continuous domain changes.

\textbf{Generalization from Reality to Artistic Styles (Non-Continuous Domain Shift):}
To assess the method’s generalization ability under non-continuous domain shifts, we follow prior works \cite{Art, Div} and formulate a cross-style transfer task. In the training phase, the model is trained on the Pascal VOC2007 and VOC2012 trainval sets, which contain natural scene images with realistic lighting, textures, and object proportions.
In the testing phase, we evaluate the model on three artistic-style datasets: Clipart1k, Watercolor2k, and Comic2k. Clipart1k includes the same 20 object classes as Pascal VOC, presented in a cartoon-like style with bold outlines and flat colors. Watercolor2k and Comic2k each contain 6 object categories, which are subsets of the Pascal VOC classes. These two datasets feature distinctive visual characteristics, soft brush strokes and color blending in Watercolor2k, and exaggerated shapes and textures in Comic2k.
In such scenarios, the domain gap between the source (real-world images) and the target (artistic-style images) is abrupt and non-continuous, with natural textures often replaced by abstract visual representations. This setup rigorously tests the method’s generalization capability under severe distributional shifts.

By addressing both continuous and non-continuous domain shifts, the two benchmark tasks provide a comprehensive framework for evaluating the robustness and adaptability of the proposed method under diverse transfer scenarios.

\subsection{Diverse Driving Weather Scenarios.}
\textbf{Daytime Clear to Day Foggy.}  In the field of autonomous driving, foggy weather is a common adverse condition that significantly reduces visibility and blurs object contours, posing major challenges for object detection tasks. Accurately identifying objects in foggy environments is critical to ensuring the safety of autonomous driving systems.
\begin{table}[!t]
  \centering
  \fontsize{8}{6}\selectfont
  \resizebox{\linewidth}{!}{
    \begin{tabular}{cc|ccccccc|c}
    \toprule
    \toprule
     \multicolumn{2}{c|}{\multirow{2}[4]{*}{Method}} & \multicolumn{7}{c|}{AP}                     & \multicolumn{1}{c}{mAP} \\
\cmidrule{3-10}    \multicolumn{2}{c|}{} & \multicolumn{1}{c|}{Bus} & \multicolumn{1}{c|}{Bike} & \multicolumn{1}{c|}{Car} & \multicolumn{1}{c|}{Motor} & \multicolumn{1}{c|}{Person} & \multicolumn{1}{c|}{Rider} & \multicolumn{1}{c|}{Truck} & \multicolumn{1}{c}{All} \\
    \midrule
    \multicolumn{2}{c|}{Faster R-CNN \cite{ren2015faster}} & 28.10  & 29.70  & 49.70  & 26.30  & 33.20  & 35.50  & 21.50  & 32.00  \\
    \multicolumn{2}{c|}{S-DGOD(Res101) \cite{S_DG}} & 32.90  & 28.00  & 48.80  & 29.80  & 32.50  & 38.20  & 24.10  & 33.50  \\
    \multicolumn{2}{c|}{C-Gap(Res101) \cite{C_Cap}} & 36.10  & 34.30  & 58.00  & 33.10  & 39.00  & 43.90  & 25.10  & 38.50  \\
    \multicolumn{2}{c|}{PDOC(Res101) \cite{pdoc}} & 36.10  & 34.50  & 58.40  & 33.30  & 40.50  & 44.20  & 26.20  & 39.10  \\
     \multicolumn{2}{c|}{UFR(Res101) \cite{UFR}} & 36.90  & 35.80  & 61.70 & 33.70 & 39.50  & 42.20  & 27.50  & 39.60  \\
      \multicolumn{2}{c|}{G-NAS(Res101) \cite{G-NAS}} &32.40 &31.20  &57.70 &31.90 &38.60 &38.50 &24.50  & 36.40 \\
      \multicolumn{2}{c|}{SECOT(Res101)\cite{SECOT}} & 39.33 & 36.28 & 60.82 & 33.79 & 39.41 & 42.68  & 31.77 & 40.58\\
       \multicolumn{2}{c|}{FWCL(Res101)\cite{FWCL}} & 31.40 & 24.90 & 50.80 & 25.70 & 32.50 & 37.50 & 23.50 & 32.30\\
    \midrule
    \multicolumn{2}{c|}{Ours(Res101)} &41.59 & 37.65 & 62.40 & 25.84 & 43.91 & 44.70 & 32.50 & 41.23 \\
    \multicolumn{2}{c|}{Ours(Swin-T)} & \textbf{45.59} & \textbf{39.21} & \textbf{64.02} & \textbf{41.06} & \textbf{47.82} & \textbf{50.32} & \textbf{37.02} & \textbf{46.40} \\

    \bottomrule
    \bottomrule
    \end{tabular}%
    }
     \caption{Per-class results (\%) on Daytime Clear to Day Foggy. \textbf{The bold sections} represent the best results.}
  \label{DF}%
\end{table}%
As shown in Table \ref{DF}, compared with other methods that rely on static target-domain text prompts (e.g., SECOT \cite{SECOT} and PDOC \cite{pdoc}), our approach consistently achieves the best performance across almost all categories when using the same backbone network. For instance, in the Person category, Ours (Res101) achieves an AP of 43.91\%, outperforming SECOT's 39.41\%. When adopting a more advanced backbone, such as the Swin Transformer \cite{swin}, our method demonstrates even greater improvements, for example, in the Bike category, our method achieves 39.21\%, significantly higher than SECOT's 36.28\%. In the Car category, our result of 64.02\% also surpasses both SECOT (60.82\%) and UFR \cite{UFR} (61.70\%).
These results verify the superiority of our LTFE method in continuous domain adaptation tasks. Through progressive feature evolution, dynamic parameter adjustment, and semantic consistency constraints, our model achieves notable improvements in object detection performance under challenging conditions like fog, thereby providing strong support for enhancing the environmental adaptability of autonomous driving systems.

\textbf{Daytime Clear to Dusk Rainy.} In autonomous driving scenarios, rainy weather during dusk presents a particularly challenging setting due to reduced illumination and visual degradation caused by raindrops and reflections on surfaces. These conditions obscure object boundaries and reduce sensor reliability, thereby increasing the difficulty of accurate object detection. Robust performance under such conditions is essential to guarantee safety in autonomous navigation.
\begin{table}[!t]
  \centering
  
   \fontsize{8}{6}\selectfont
  \resizebox{\linewidth}{!}{
    \begin{tabular}{cc|ccccccc|c}
    \toprule
    \toprule
    \multicolumn{2}{c|}{\multirow{2}[4]{*}{Method}} & \multicolumn{7}{c|}{AP}                     & \multicolumn{1}{c}{mAP} \\
\cmidrule{3-10}    \multicolumn{2}{c|}{} & \multicolumn{1}{c|}{Bus} & \multicolumn{1}{c|}{Bike} & \multicolumn{1}{c|}{Car} & \multicolumn{1}{c|}{Motor} & \multicolumn{1}{c|}{Person} & \multicolumn{1}{c|}{Rider} & \multicolumn{1}{c|}{Truck} & \multicolumn{1}{c}{All} \\
    \midrule
    \multicolumn{2}{c|}{Faster R-CNN \cite{ren2015faster}} & 28.50  & 20.30  & 58.20  & 6.50  & 23.40  & 11.30  & 33.90  & 26.00  \\
    \multicolumn{2}{c|}{S-DGOD \cite{S_DG}} & 37.10  & 19.60  & 50.90  & 13.40  & 19.70  & 16.30  & 40.70  & 28.20  \\
    \multicolumn{2}{c|}{C-Gap \cite{C_Cap}} & 37.80  & 22.80  & 60.70  & 16.80  & 26.80  & 18.70  & 42.40  & 32.30  \\
    \multicolumn{2}{c|}{PDOC \cite{pdoc}} & 39.40  & 25.20  & 60.90  & 20.40  & 29.90  & 16.50  & 43.90  & 33.70  \\
    \multicolumn{2}{c|}{UFR \cite{UFR}} & 37.10  & 21.80  & 67.90 & 16.40  & 27.40  & 17.90  & 43.90  & 33.20  \\
   \multicolumn{2}{c|}{G-NAS(Res101) \cite{G-NAS}} &44.60 & 22.30 & 66.40 & 14.70 & 32.10 & 19.60 & 45.80 &35.10 \\
      \multicolumn{2}{c|}{SECOT(Res101)\cite{SECOT}} & 45.75 & 19.20 & \textbf{74.32} & 23.43 & 35.81 & 22.14 & 53.47 & 39.16 \\
       \multicolumn{2}{c|}{FWCL(Res101)\cite{FWCL}} & 37.20 & 22.60 & 60.50 & 17.00 & 27.00 & 19.90 & 43.80 & 32.60\\
    \midrule
     \multicolumn{2}{c|}{Ours(Res101)} & \textbf{49.31} & 25.89 & 60.56 & 20.34 & 45.20 & 24.31 & 52.36 & 39.71 \\
    \multicolumn{2}{c|}{Ours(Swin-T)} & 46.18 & \textbf{42.71} & 65.12 & \textbf{35.82} & \textbf{48.18} & \textbf{33.40} & \textbf{54.35} & \textbf{46.54}\\

    \bottomrule
    \bottomrule
    \end{tabular}%
    }
    \caption{Per-class results (\%) on Daytime Clear to Dusk Rainy. \textbf{The bold} represent the best results.}
  \label{DR}%
  
\end{table}%
As shown in Table \ref{DR}, our method significantly outperforms existing approaches across most categories. Compared with SECOT \cite{SECOT} and PDOC \cite{pdoc}, our approach demonstrates superior adaptability under adverse rainy conditions. For instance, when using ResNet-101 as the backbone, our method achieves an AP of 49.31\% on the Bus category, surpassing SECOT (45.75\%) and PDOC (39.40\%). In the Person category, our method obtains 45.20\%, significantly outperforming SECOT (35.81\%) and UFR \cite{UFR} (27.40\%). Furthermore, in the Rider category, our approach achieves 24.31\%, outperforming both SECOT (22.14\%) and PDOC (16.50\%).
When equipped with a more advanced backbone, Swin Transformer \cite{swin}, our method further boosts performance across all categories. Notably, in the Bike category, it reaches 42.71\%, significantly ahead of SECOT (19.20\%) and PDOC (25.20\%); in Motor, we observe 35.82\% compared to SECOT’s 23.43\% and PDOC’s 20.40\%. Across all categories, our method achieves the highest overall mAP of 46.54\%, clearly outperforming previous state-of-the-art methods such as SECOT (39.16\%) and G-NAS \cite{G-NAS} (35.10\%).

These results strongly validate the effectiveness of our LTFE framework in handling continuous domain shifts under complex weather and lighting variations. Through liquid temporal feature evolution, dynamic adjustment of parameters, and semantic consistency regularization, our model delivers robust detection performance in dusk-to-rainy transitions, further improving the resilience and reliability of autonomous systems in real-world conditions.

\textbf{Daytime Clear to Night Sunny.} In autonomous driving scenarios, domain adaptation from nighttime to bright daytime conditions presents a particularly challenging task. Night scenes typically suffer from low visibility, motion blur, and strong lighting contrasts, whereas sunny daytime conditions introduce shadows, glare, and more complex background distractions. These significant domain shifts can severely hinder the generalization performance of object detection models.
As shown in Table \ref{NS}, our method demonstrates strong robustness and adaptability across multiple critical categories, clearly outperforming mainstream baselines. For example, using ResNet-101 as the backbone, our model achieves an AP of 50.76\% in the Bus category, which is 8.81\% higher than SECOT \cite{SECOT} (41.95\%) and 7.16 points higher than UFR \cite{UFR}(43.60\%).
When employing the more advanced Swin Transformer as the backbone, our method achieves even greater improvements. In the Bike category, we reach 51.37\%, representing an increase of 14.16 percentage points over SECOT (37.21\%). In the Motor category, our model achieves 33.60\%, surpassing G-NAS (26.50\%) and PDOC (21.30\%) by 7.10\% and 12.30\%, respectively. In terms of overall performance, our approach achieves a mAP of 53.14\% across all categories, exceeding SECOT (42.00\%) by 11.14 percentage points and G-NAS (45.00\%) by 8.14 points.
These results further validate the effectiveness of our proposed LTFE framework in handling severe lighting changes and structural variations between scenes. Through continuous feature evolution modeling, dynamic parameter modulation, and semantic consistency constraints, our model demonstrates robust object detection capabilities in the challenging Night-to-Sunny domain shift, offering improved reliability and adaptability for real-world autonomous driving systems.

\begin{table}[t]
  \centering

 \fontsize{8}{6}\selectfont
  \resizebox{\linewidth}{!}{
    \begin{tabular}{cc|ccccccc|c}
    \toprule
    \toprule
   \multicolumn{2}{c|}{\multirow{2}[4]{*}{Method}} & \multicolumn{7}{c|}{AP}                     & \multicolumn{1}{c}{mAP} \\
\cmidrule{3-10}    \multicolumn{2}{c|}{} & \multicolumn{1}{c|}{Bus} & \multicolumn{1}{c|}{Bike} & \multicolumn{1}{c|}{Car} & \multicolumn{1}{c|}{Motor} & \multicolumn{1}{c|}{Person} & \multicolumn{1}{c|}{Rider} & \multicolumn{1}{c|}{Truck} & \multicolumn{1}{c}{All} \\
    \midrule
    \multicolumn{2}{c|}{Faster R-CNN \cite{ren2015faster}} & 34.70  & 32.00  & 56.60  & 13.60  & 37.40  & 27.60  & 38.60  & 34.40  \\
    \multicolumn{2}{c|}{S-DGOD \cite{S_DG}} & 40.60  &35.10  & 50.70  & 19.70  & 34.70  & 32.10  & 43.40  & 36.60  \\
    \multicolumn{2}{c|}{C-Gap \cite{C_Cap}} & 37.70  & 34.30  & 58.00  & 19.20  & 37.60  & 28.50  & 42.90  & 36.90  \\
    \multicolumn{2}{c|}{PDOC \cite{pdoc}} & 40.90  & 35.00  & 59.00  & 21.30  & 40.40  & 29.90 & 42.90  & 38.50  \\
    \multicolumn{2}{c|}{UFR \cite{UFR}} & 43.60  & 38.10  & 66.10 & 14.70  & 49.10  & 26.40  & 47.50  & 40.80  \\
    \multicolumn{2}{c|}{G-NAS(Res101) \cite{G-NAS}} &46.90 & 40.50 & 67.50 & 26.50 & 50.70 & 35.40 & 47.80 & 45.00 \\
      \multicolumn{2}{c|}{SECOT(Res101)\cite{SECOT}} & 41.95 & 37.21 & 66.19 & 21.56 & 49.61 & 33.25 & 44.24 & 42.00 \\
       \multicolumn{2}{c|}{FWCL(Res101)\cite{FWCL}} &39.60 & 36.40 & 58.90 & 18.00 & 42.60 & 26.20 & 40.50 & 37.50 \\
    \midrule
    
    \multicolumn{2}{c|}{Ours(Res101)} & 50.76 & 35.74 & 61.27 & 23.54 & 50.29 & 32.56 & 47.21 & 43.05 \\
\multicolumn{2}{c|}{Ours(Swin-T)} & \textbf{52.54} & \textbf{51.37} & \textbf{71.02} & \textbf{33.60} & \textbf{60.25} & \textbf{43.87} & \textbf{59.36} & \textbf{53.14}
 \\
    
    \bottomrule
    \bottomrule
    \end{tabular}%
    }
     \caption{Per-class results (\%) on Daytime Clear to Night Sunny. \textbf{The bold} represent the best results.}
  \label{NS}%
\end{table}%

\textbf{Daytime Clear to Night Rainy.} In single-source domain generalization tasks, transferring from daytime clear conditions to nighttime rainy environments presents a highly challenging scenario. The domain gap in this setting arises not only from drastic changes in illumination but also from adverse effects introduced by rain, such as motion blur, occlusions, water streaks, and significant degradation of object appearance information. These factors severely interfere with object boundaries and semantic cues, thereby increasing the difficulty of accurate detection in rainy nighttime scenes.
As shown in Table \ref{NR}, our method achieves leading performance across several critical object categories. With ResNet-101 as the backbone, we obtain an Average Precision (AP) of 20.56\% for the Person category and 17.25\% for Rider, both outperforming SECOT’s results of 16.67\% and 14.20\%, respectively.
When employing a more advanced backbone such as Swin Transformer, our approach exhibits notable improvements across all categories. For instance, in the Car category, our method achieves 55.78\% AP, surpassing SECOT’s 42.59\% by a margin of 13.19 percentage points. For the Person and Rider categories, we reach 30.61\% and 22.83\% AP, significantly outperforming all previous methods. In the large vehicle categories of Bus and Truck, we achieve 45.91\% and 47.67\% AP, notably exceeding SECOT’s 37.85\% and 30.55\%.

Overall, our method achieves a mean Average Precision (mAP) of 33.40\% in the Night Rainy setting, demonstrating the robustness and strong cross-domain generalization capabilities of our proposed LTFE method. By modeling temporal feature evolution and introducing a semantic alignment mechanism, our model effectively adapts to the image degradation and structural ambiguities caused by rain, offering a more reliable perception framework for autonomous driving in complex, low-light, rainy environments.

\begin{table}[!t]
  \centering

   \fontsize{8}{6}\selectfont
  \resizebox{\linewidth}{!}{
    \begin{tabular}{cc|ccccccc|c}
    \toprule
    \toprule
    \multicolumn{2}{c|}{\multirow{2}[4]{*}{Method}} & \multicolumn{7}{c|}{AP}                     & \multicolumn{1}{c}{mAP} \\
\cmidrule{3-10}    \multicolumn{2}{c|}{} & \multicolumn{1}{c|}{Bus} & \multicolumn{1}{c|}{Bike} & \multicolumn{1}{c|}{Car} & \multicolumn{1}{c|}{Motor} & \multicolumn{1}{c|}{Person} & \multicolumn{1}{c|}{Rider} & \multicolumn{1}{c|}{Truck} & \multicolumn{1}{c}{All} \\
    \midrule
    \multicolumn{2}{c|}{Faster R-CNN \cite{ren2015faster}} & 16.80  & 6.90  & 26.30  & 0.60  & 11.60  & 9.40  & 15.40  & 12.40  \\
    \multicolumn{2}{c|}{S-DGOD \cite{S_DG}} & 24.40  & 11.60  & 29.50  & 9.80  & 10.50  &11.40  & 19.20  & 16.60  \\
    \multicolumn{2}{c|}{C-Gap \cite{C_Cap}} & 28.60  & 12.10  & 36.10  & 9.20  & 12.30  & 9.60  & 22.90  & 18.70  \\
    \multicolumn{2}{c|}{PDOC \cite{pdoc}} & 25.60  & 12.10  & 35.80  & 10.10  & 14.20  & 12.90 & 22.90  & 19.20  \\
    \multicolumn{2}{c|}{UFR \cite{UFR}} & 29.90 & 11.80  & 36.10 & 6.40  & 13.10  & 10.50  & 23.30  & 19.20  \\
    \multicolumn{2}{c|}{G-NAS(Res101) \cite{G-NAS}} &28.60 & 9.80 & 38.40 & 0.10 & 13.80 & 9.80 & 21.40& 17.40 \\
      \multicolumn{2}{c|}{SECOT(Res101)\cite{SECOT}} & 37.85 & 16.98 & 42.59 & \textbf{12.89} & 16.67 & 14.20 & 30.55 & 24.53 \\
       \multicolumn{2}{c|}{FWCL(Res101)\cite{FWCL}} &28.90 & 11.20 & 34.60 & 9.10 & 12.10 & 12.00 & 23.90 & 18.90 \\
    \midrule
    
    \multicolumn{2}{c|}{Ours(Res101)} & 34.82 & 16.71 & 38.41 & 11.60 & 20.56 & 17.25 & 30.73 & 24.30 \\
    \multicolumn{2}{c|}{Ours(Swin-T)} & \textbf{45.91} & \textbf{18.64} & \textbf{55.78} & 12.37 & \textbf{30.61} & \textbf{22.83} & \textbf{47.67} & \textbf{33.40} \\

    \bottomrule
    \bottomrule
    \end{tabular}%
    }
\caption{Per-class results (\%) on Daytime Clear to Night Rainy. \textbf{The bold} represent the best results.}
  \label{NR}%
\end{table}%

\begin{table}[h]
  \centering
  
   \fontsize{8}{10}\selectfont
  \resizebox{\linewidth}{!}{
    \begin{tabular}{cc|cccccc|c}
    \toprule
    \toprule
    \multicolumn{2}{c|}{\multirow{2}[4]{*}{Method}} & \multicolumn{6}{c|}{AP}               & \multicolumn{1}{c}{mAP} \\
\cmidrule{3-9}    \multicolumn{2}{c|}{} & \multicolumn{1}{c|}{Bike} & \multicolumn{1}{c|}{Bird} & \multicolumn{1}{c|}{Car} & \multicolumn{1}{c|}{Cat} & \multicolumn{1}{c|}{Dog} & \multicolumn{1}{c|}{Person} & \multicolumn{1}{c}{All} \\
    \midrule
    \multicolumn{2}{c|}{Faster R-CNN\cite{ren2015faster}} & 87.60  & 41.60  & 36.40  & 29.60  & 18.70  & 54.50  & 44.63  \\
    \multicolumn{2}{c|}{NP(Res101)\cite{np}} & 87.31  & 56.22  & 50.37  & 42.05  & 41.78  & 42.18  & 53.31  \\
    \multicolumn{2}{c|}{C-Gap(Res101)\cite{C_Cap}} & 86.51  & 52.59  & 49.10  & 40.55  & 39.57  & 35.68  & 50.68  \\
   \multicolumn{2}{c|}{DIV(Res101)\cite{Div}} & 90.40  & 51.80  & 51.90  & 43.90  & 35.90  & \textbf{70.20}  & 57.40  \\
    \multicolumn{2}{c|}{SECOT(Res101)\cite{SECOT}} & 91.88 & 56.74 & 50.41 & \textbf{48.22} & 41.57 & 56.19 & 57.50  \\
    \midrule
    \multicolumn{2}{c|}{Ours(Res101)} & 92.40 & 57.13 & 52.64 & 47.18 & 42.69 & 57.40 & 58.24 \\
    \multicolumn{2}{c|}{Ours(Swin-T)} & \textbf{92.70} & \textbf{63.18} & \textbf{55.47} & 46.31 & \textbf{46.51} & 63.70 & \textbf{61.31} \\
    
    \bottomrule
    \bottomrule
    \end{tabular}%
    }
  \caption{Per-class results (\%) on VOC to Watercolor. \textbf{The bold sections} represent the best results.}
   \label{watecolor}
\end{table}%

\subsection{Generalization from Reality to Art.}

To evaluate the generalization capability of our proposed method across different types of data domains, we focus on the challenging non-continuous domain shift between real and artistic domains, such as the transfer from PASCAL VOC to the Clipart, Watercolor, and Comic datasets, which exhibit significant and complex distribution differences. Table \ref{watecolor} presents the generalization results from PASCAL VOC to the Watercolor dataset. The experiments show that under the same backbone network (Res101), our method outperforms other baselines and existing generalization approaches across nearly all categories.
Specifically, our method achieves high detection accuracy in almost all categories, with the most notable improvement in the Person category, reaching 57.40\%, surpassing SECOT’s 56.19\%. When adopting the more advanced Swin Transformer backbone, performance further improves, with an overall mAP of 61.31\%, fully demonstrating the strong adaptability of our method to non-continuous domain shifts.

Tables \ref{comic} and \ref{clipart} show the generalization results from PASCAL VOC to the Comic and Clipart datasets, respectively. These tables indicate that using the same Res101 backbone, our method achieves further performance gains on these two datasets, which also represent non-continuous domain shifts with significant style differences, validating the generalization capability of our method across various style target domains.
These results demonstrate that by introducing temporal perturbations to simulate feature evolution, we can effectively model large domain shifts. This approach applies not only to continuous domain biases but also to non-continuous ones, significantly enhancing the model’s generalization performance under complex distribution differences between real and artistic domains, highlighting the superiority of our method in addressing non-continuous domain adaptation tasks.

\begin{table}[t]
  \centering
 
   \fontsize{8}{6}\selectfont
  \resizebox{\linewidth}{!}{
    \begin{tabular}{cc|cccccc|c}
    \toprule
    \toprule
    \multicolumn{2}{c|}{\multirow{2}[4]{*}{Method}} & \multicolumn{6}{c|}{AP}               & \multicolumn{1}{c}{mAP} \\
\cmidrule{3-9}    \multicolumn{2}{c|}{} & \multicolumn{1}{c|}{Bike} & \multicolumn{1}{c|}{Bird} & \multicolumn{1}{c|}{Car} & \multicolumn{1}{c|}{Cat} & \multicolumn{1}{c|}{Dog} & \multicolumn{1}{c|}{Person} & \multicolumn{1}{c}{All} \\
    \midrule
    \multicolumn{2}{c|}{Faster R-CNN\cite{ren2015faster}} & 36.50  & 8.60  & 25.90  & 9.20  & 10.80  & 25.20  & 19.37  \\
    \multicolumn{2}{c|}{NP\cite{np}} & 42.44   & 18.25  & 38.79  & 17.33  & 24.29  & 32.18  & 28.88  \\
    \multicolumn{2}{c|}{C-Gap\cite{C_Cap}} & 41.98  & 15.94  & 41.97  & 18.77  & 24.60  & 33.26 & 29.42 \\
    \multicolumn{2}{c|}{SECOT\cite{SECOT}} & 47.00 & 22.99 & 44.95 & 20.91 & 29.42 & 43.63 & 34.82  \\
    \midrule
    \multicolumn{2}{c|}{Ours(Res101)} & 41.29 & \textbf{23.49} & 43.17 & 22.83 & \textbf{33.74} & 47.60 & 35.35  \\
    \multicolumn{2}{c|}{Ours(Swin)} & \textbf{51.45} & 22.31 & \textbf{47.60} & \textbf{22.95} & 30.64 & \textbf{48.90} & \textbf{37.31} \\
    \bottomrule
    \bottomrule
    \end{tabular}%
    }
     \caption{Per-class results (\%) on VOC to Comic. \textbf{The bold sections} represent the best results.}
  \label{comic}%
\end{table}%

\begin{table}[!h]
  \centering
 
   \fontsize{12}{13}\selectfont
   \resizebox{1.0\linewidth}{!}{
    \begin{tabular}{ccrc|cc}
    \toprule
    \toprule
    \multicolumn{1}{{c}}{Layer1} & \multicolumn{1}{{c}}{Layer2} & \multicolumn{1}{{c}}{Layer3} & \multicolumn{1}{{c}|}{Layer4} & \multicolumn{1}{{c}}{Night Sunny} & \multicolumn{1}{{c}}{Night Rainy}\\
    \midrule
     $\checkmark$     &       &       &       & \textbf{43.1} & \textbf{24.3} \\
          &  $\checkmark$     &       &       & 40.5 & 24.1 \\
          &       & $\checkmark$ & &  41.6     & 21.5  \\
          &       &       &  $\checkmark$     & 40.2 & 20.3 \\
    \bottomrule
    \bottomrule
    \end{tabular}%
    }
     \caption{Ablation study on the impact of selecting different feature layers for Progressive Temporal Feature Evolution. ‘Layer1’ denotes applying Progressive Temporal Feature Evolution starting from the first feature layer. \textbf{The bold sections} represent the best results.}
  \label{FS}%
\end{table}%

\begin{table*}[!t]
  \centering  
  \resizebox{\textwidth}{!}{%
    \begin{tabular}{cc|cccccccccccccccccccc|c}
    \toprule
    \toprule
    \multicolumn{2}{c|}{\multirow{2}[4]{*}{Method}} & \multicolumn{20}{c|}{AP}                                                                                                                              & \multicolumn{1}{c}{mAP} \\
\cmidrule{3-23}    \multicolumn{2}{c|}{} & \multicolumn{1}{c}{Place} & \multicolumn{1}{c}{Bike} & \multicolumn{1}{c}{Bird} & \multicolumn{1}{c}{Boat} & \multicolumn{1}{c}{Bottle} & \multicolumn{1}{c}{Bus} & \multicolumn{1}{c}{Car } & \multicolumn{1}{c}{Cat} & \multicolumn{1}{c}{Chair} & \multicolumn{1}{c}{Cow} & \multicolumn{1}{c}{Table} & \multicolumn{1}{c}{Dog} & \multicolumn{1}{c}{Horse} & \multicolumn{1}{c}{Mot.} & \multicolumn{1}{c}{Person} & \multicolumn{1}{c}{Plant} & \multicolumn{1}{c}{Sheep} & \multicolumn{1}{c}{Sofa} & \multicolumn{1}{c}{Train} & \multicolumn{1}{c|}{TV} & \multicolumn{1}{c}{All} \\
    \midrule
    \multicolumn{2}{c|}{Faster R-CNN\cite{ren2015faster}} & 34.63  & 17.02  & 28.39  & 16.37  & 13.90  & 35.51  & 44.64  & 35.14  & 19.60  & 29.05  & 20.48  & 25.70  & 39.43  & 24.41  & 34.87  & 8.98  & 17.03  & 27.78  & 29.73  & 26.96  & 26.50  \\
    \multicolumn{2}{c|}{NP\cite{np}} & 40.91  & 44.30  & 38.02  & 24.26  & 26.78  & 47.04  & 56.65  & 33.80  & 35.84  & 31.53  & 32.40  & 26.20  & 45.73  & 38.79  & 41.64  & 10.39  & 15.97  & \textbf{41.69}  & 38.17  & 37.85  & 35.40  \\
    \multicolumn{2}{c|}{C-Gap\cite{C_Cap}} & 46.08  & 42.18  & 41.04  & 24.86  & 27.25  & 43.40  & 54.60  & 39.81  & 33.27  & 40.06  & 28.50  & 32.95  & 55.32  & 38.95  & 45.03  & 10.66  & 21.54  & 37.31  & 32.68  & 38.43  & 36.74  \\
     \multicolumn{2}{c|}{SECOT\cite{SECOT}} & 51.22 & 51.86 & \textbf{50.24} & 28.93 & 26.99 & 46.23 & 55.19 & 44.46 & 33.61 & 38.26 & 34.66 & 40.22 & \textbf{59.82} & 42.82 & 49.12 & 13.13 & 25.93 & 34.75 & 37.87 & 38.29 & 40.20 \\
    \midrule
    \multicolumn{2}{c|}{Ours(Res101)} & 51.74 & 46.85 & 40.72 & 36.86 & 29.43 & 44.76 & 60.96 & \textbf{54.32} & 32.47 & 44.65 & 35.07 & 42.13 & 55.68 & \textbf{43.17} & 49.24 & 25.33 & 27.30 & 36.87 & 42.93 & 41.24 & 42.09  \\
    \multicolumn{2}{c|}{Ours(Swin)} & \textbf{52.74} & \textbf{53.20} & 49.41 & \textbf{40.39} & \textbf{27.60} & \textbf{52.73} & \textbf{62.37} & 42.80 & \textbf{42.14} & \textbf{45.53} & \textbf{39.04} & \textbf{43.78} & 50.37 & 41.26 & \textbf{56.40} & \textbf{26.78} & \textbf{29.30} & 41.65 & \textbf{46.79} & \textbf{45.86} & \textbf{44.51} \\
    \bottomrule
    \bottomrule
    \end{tabular}%
    }
    \caption{Per-class results (\%) on VOC to Clipart. \textbf{The bold sections} represent the best results.}
  \label{clipart}%
\end{table*}

\begin{table}[!h]
  \centering
 
   \fontsize{12}{13}\selectfont
   \resizebox{1.0\linewidth}{!}{
    \begin{tabular}{ccc|ccc}
    \toprule
    \toprule
    \multicolumn{1}{{c}}{One-Shot} & \multicolumn{1}{{c}}{Equal-Step} & \multicolumn{1}{{c}|}{Progressive} & \multicolumn{1}{{c}}{Night Sunny } & \multicolumn{1}{{c}}{Day
    Foggy} & \multicolumn{1}{{c}}{Night Rainy}\\
    \midrule
     $\checkmark$     &       &       &   37.6    & 38.4 & 16.7 \\
          &  $\checkmark$     &       &   41.7    & 40.1 & 20.6 \\
          &       & $\checkmark$ &  \textbf{43.1}  &  \textbf{41.2}     & \textbf{24.3}  \\
    \bottomrule
    \bottomrule
    \end{tabular}%
    }
     \caption{Ablation study on the noise and blur injection strategy. ‘One-Shot’ refers to injecting the maximum intensity of noise and blur in a single step.
    ‘Equal-Step’ denotes injecting a fixed level of noise and blur at each iteration.
    ‘Progressive’, which is the strategy adopted in our method, involves gradually injecting noise and blur in a stepwise manner.}
  \label{NJ}%
\end{table}%

\section{Extended Ablation Experiments.}

\subsection{Impacts of Noise and Blur Injection Strategies.}
The core idea of the proposed Progressive Temporal Feature Evolution (PTFE) lies in simulating continuous domain shifts, such as the gradual transition from clear to rainy weather, by dynamically adjusting Gaussian noise (with exponentially decaying intensity $\alpha_t$) and Gaussian blur (with exponentially increasing intensity $\sigma_t$). Existing methods, like discrete data augmentation, typically apply static or one-shot perturbations that fail to capture the dynamic evolution of feature distributions. To verify the advantage of the progressive perturbation over equal-step and one-shot injection strategies, we designed ablation experiments that demonstrate its critical role in enhancing model generalization. Table \ref{NJ} compares these three noise and blur injection strategies across three scenarios: Night Sunny, Day Foggy, and Night Rainy, illustrating their impact on detection performance.

In the Night Sunny scenario, lighting is low but relatively stable, with the transition from day to night being a gradual process. Our progressive strategy (43.1\%) outperforms both one-shot injection (37.6\%) and equal-step injection (41.7\%). This is because the progressive approach simulates the slow temporal change of light scattering by gradually increasing blur intensity and exponentially decaying noise intensity, allowing the model to learn robust representations that handle gradual feature degradation in low-light environments.

The Day Foggy scenario is characterized by dynamically fluctuating fog density and progressively changing visibility rather than abrupt shifts. Here, the progressive strategy (41.2\%) also shows clear advantages: the blur intensity grows exponentially ($\sigma_t = \sigma_0 \cdot \gamma^t$), effectively modeling the gradual blurring of visual features as fog thickens, while the noise intensity decays exponentially ($\alpha_t = \alpha_0 \cdot \exp(-\lambda t)$) to avoid excessive noise interference. In contrast, one-shot injection disrupts the continuity of foggy features with sudden strong perturbations, and equal-step injection fails to capture the dynamic fog density changes due to fixed intensity levels.
The performance gap is most pronounced in the Night Rainy scenario: the progressive strategy (24.3\%) significantly outperforms equal-step injection (20.6\%) and one-shot injection (16.7\%). Rainy nights involve continuously evolving factors such as raindrop density, motion blur, and light scattering. The blur gradually intensifies as rain strengthens, while noise stabilizes after initial disturbance. The progressive approach’s dynamic parameter adjustment perfectly matches this nonlinear evolution, enabling the model to effectively distinguish raindrop noise from genuine object features. The other two strategies, lacking this temporal continuity, fall short in capturing the complex rainy night characteristics.

In summary, these results validate the necessity of the progressive noise and blur injection strategy. By simulating the continuous evolution of features under real-world weather conditions, it significantly enhances the model’s sensitivity to fine-grained cross-domain variations, providing essential support for S-DGOD tasks.

\subsection{Impacts of Feature Layers.}
To validate the influence of feature layer selection on the effectiveness of Progressive Temporal Feature Evolution (PTFE), we perform an ablation analysis. Table \ref{FS} presents the detection performance (mAP\%) of the model on two target domains, Night Sunny and Night Rainy, when applying Progressive Temporal Feature Evolution (PTFE) to different feature layers. The results demonstrate that the choice of feature layer has a significant impact on the model’s generalization ability. Specifically, applying PTFE to the first feature layer yields the highest mAP on both Night Sunny (43.1\%) and Night Rainy (24.3\%), outperforming all other settings. This suggests that low-level features, such as edges and textures, are more suitable for modeling continuous domain shifts (e.g., visual degradation caused by weather changes), effectively bridging the distribution gap between the source and target domains.

The application of PTFE to the second layer results in slightly lower performance than the first layer, but it remains superior to the higher layers. This could be attributed to the second layer still retaining some low-level visual cues, although its higher abstraction level reduces sensitivity to fine-grained visual changes. In contrast, applying PTFE to the Layer 3 and Layer 4 layers leads to a noticeable performance drop. For instance, using the fourth layer results in only 20.3\% mAP under the Night Rainy condition, significantly lower than the 24.3\% achieved by the first layer. This is likely because high-level features are more semantically oriented and less capable of capturing low-level visual variations such as those introduced by fog or rain, thereby weakening the model's ability to simulate continuous domain transitions.

\subsection{Component Analysis.}
To validate the effectiveness of each core component in our proposed model, we conducted a comprehensive ablation study on Progressive Temporal Feature Evolution (PTFE), the combined module of Temporal Dependency Modeling and Liquid Parameter Evolution (TDM+LPE), and the Temporal Feature Alignment Module (TFAM). Since LPE relies on the sequential information provided by TDM for dynamic adjustment, the two components are evaluated jointly. The experimental results are summarized in Table \ref{CA}, where the first row corresponds to the Faster R-CNN baseline, and subsequent rows show the performance after progressively integrating each module.

Without any of the proposed components, the baseline model achieves an mAP of 49.2\% on the source domain (e.g., Sunny Day), but performs poorly on the target domains (e.g., Night Sunny, Dusk Rainy), with an mAP as low as 13.7\% under Night Rainy conditions. This indicates a clear struggle in adapting to the domain shift between source and target distributions.
When PTFE is introduced, the model exhibits an initial performance improvement on the target domains, for example, the mAP on Night Sunny increases from 35.4\% to 36.8\%. This suggests that PTFE, by simulating the initial feature evolution trajectory from the source to the latent target domain, can enhance the model’s ability to capture feature diversity. However, the absence of dynamic optimization mechanisms limits the extent of improvement.

The third row in the table shows the results after further incorporating TDM+LPE on top of PTFE. Here, the model achieves substantial gains, for instance, the mAP under Dusk Rainy increases from 29.3\% to 39.4\%, and Night Rainy improves from 14.1\% to 22.7\%. This indicates that TDM, through LSTM, effectively captures the temporal dependencies in feature evolution, while LPE leverages neural ODEs to generate dynamic convolutional kernels, enabling more precise control over the evolution trajectory. Together, they significantly enhance detection performance under unseen target domains.

The fourth row presents the performance of adding TFAM to PTFE. While this combination outperforms PTFE alone, for example, Night Sunny mAP increases from 36.8\% to 40.2\% and Night Rainy from 14.1\% to 19.7\%, it still falls short of the PTFE+TDM+LPE combination. This result demonstrates that TFAM, by enforcing intra-class consistency and inter-class separability during feature evolution, helps reduce the loss of semantic target information. However, the lack of dynamic parameter adaptation continues to limit its performance ceiling.

The final row of the table reports the performance of our full model, with PTFE, TDM+LPE, and TFAM integrated. In this setting, the model achieves the best performance across all scenarios. These results validate the complementary nature of the components: PTFE provides the initial evolution trajectory, TDM+LPE enables dynamic modeling and adaptive evolution, and TFAM ensures semantic consistency. Together, they produce a more realistic simulation of cross-domain feature transitions and significantly enhance the model's generalization capability.

\begin{table}[!t]
  \centering

  \fontsize{22}{22}\selectfont
     \resizebox{\linewidth}{!}{
    \begin{tabular}{ccc|c|cccc}
    \toprule
    \toprule
    \multicolumn{3}{c|}{Method} & \multicolumn{1}{c|}{Source} & \multicolumn{4}{c}{Target} \\
    \midrule
    \multicolumn{1}{c}{PTFE} & \multicolumn{1}{c}{TDM+LPE}  & \multicolumn{1}{c|}{TFAM} & \multicolumn{1}{c|}{Day Clear} &  \multicolumn{1}{c}{Night Sunny}& \multicolumn{1}{c}{Dusk Rainy} & \multicolumn{1}{c}{Night Rainy}  & \multicolumn{1}{c}{Day Foggy}  \\
    
    \midrule
         &        &       &  49.2  & 35.4  & 27.5  & 13.7  & 32.6\\
    $\checkmark$     &           &       & 54.6  & 36.8  & 29.3  & 14.1  & 33.6 \\
    $\checkmark$     &  $\checkmark$         &       & 57.3  & 42.6  & 39.4  & 22.7  & 40.8  \\  
    
    $\checkmark$     &      &  $\checkmark$     & 56.8  & 40.2  & 34.1  & 19.7  & 36.5 \\
    $\checkmark$     & $\checkmark$       & $\checkmark$     & \textbf{58.4} & \textbf{43.1} & \textbf{39.7} & \textbf{24.3} & \textbf{41.2} \\
    \bottomrule
    \bottomrule
    \end{tabular}%
    }
  
   \caption{\textbf{Ablation analysis of our proposed model.} PTFE stands for  Progressive Temporal Feature Evolution, TDM + LPE denotes the combination of Temporal Dependency Modeling and Liquid Parameter Evolution, and TFAM denotes Temporal Feature Alignment Module.}
   \label{CA}%
\end{table}%

\section{Visualization Analysis}
\begin{figure*}[!h]
  \centering
  \includegraphics[width=2.0\columnwidth]{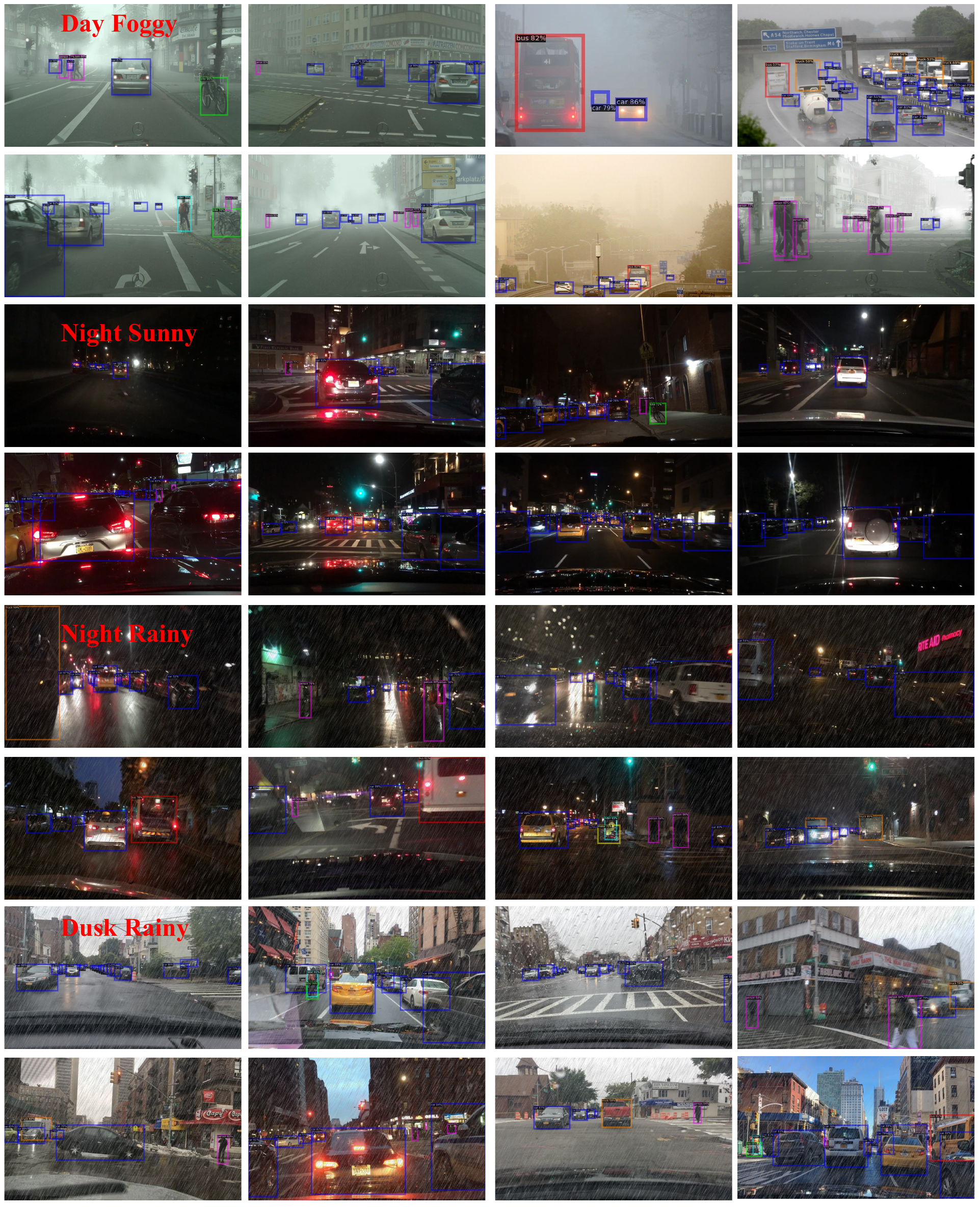}
  \caption{\textbf{Visualization analysis of our method.} The figure illustrates the detection results of our method under four different weather conditions.}
    \label{Mvis}
\end{figure*}

To further validate the effectiveness of our method, we conduct a visualization analysis of detection results under four challenging weather conditions: Day Foggy, Night Sunny, Night Rainy, and Dusk Rainy, as shown in Figure \ref{Mvis}.

The first and second rows correspond to the Day Foggy condition, where the primary challenge lies in low visibility and blurred object edges, resulting in a continuous degradation of visual features from clarity to obscurity. The visualization shows that our model can accurately detect vehicles, pedestrians, and other objects at varying distances, even in dense fog regions, with a low miss rate. This is attributed to our use of controllable Gaussian blur to simulate the progressive increase in fog density, enabling the model to learn the continuous feature transition pattern from ‘clear’ to ‘light blur’ to ‘heavy blur’.

The third and fourth rows present results under the Night Sunny condition. The key difficulty here is the combination of low and uneven lighting, such as the contrast between streetlights and shadowed areas, which can cause significant feature distortion. The visualizations demonstrate that our model outperforms competing methods in recognizing small objects such as pedestrians and bicycles in low-light environments, with notably better localization in shadowed areas. This performance stems from our temporal modeling capability, which captures the temporal correlations of lighting changes. In addition, the liquid neural network adaptively adjusts evolving features in real time, enhancing edge responses in dark areas while suppressing noise in overexposed regions.

The fifth and sixth rows illustrate the Night Rainy scenario, widely regarded as one of the most difficult environments for autonomous driving. This setting involves multiple interfering factors, including low light, raindrop noise, and object reflections, which can lead to feature confusion, such as mistaking water reflections for vehicle lights. The visualizations reveal that our model can effectively distinguish raindrop noise from real objects and achieves high detection accuracy for large targets like cars and buses.

The last two rows show the Dusk Rainy condition, characterized by rapidly changing lighting (from dusk to night) and rainfall interference. This represents a typical ‘continuous domain shift’ scenario. The visualization indicates that our model maintains stable detection performance throughout the transition from dusk to night, with particularly high accuracy in small object detection. This demonstrates the strength of our method in modeling complete feature evolution trajectories. By simulating the full scene transition, from moderate lighting and light rain to low lighting and heavy rain, the model learns the cross-domain correlation between lighting intensity, rainfall conditions, and object features. As a result, it can quickly adapt to dynamic real-world changes with only a small amount of feature evolution during inference.

\clearpage

\end{document}